\documentclass[nohyperref]{article}

\usepackage[accepted]{icml2022}
\usepackage{times}
\usepackage{amsmath}
\usepackage{algorithm}

\makeatletter
\let\algorithmic\@undefined
\let\comment\@undefined
\let\endalgorithmic\@undefined    
\makeatother
\usepackage{algpseudocode}

\usepackage[utf8]{inputenc} \usepackage[T1]{fontenc}    \usepackage{hyperref}
\hypersetup{
  colorlinks=true,
  citecolor=cyan,
  linkcolor=cyan,
}
\usepackage{url}            \usepackage{booktabs}       \usepackage{amsfonts}       \usepackage{nicefrac}       
\usepackage{microtype}      \usepackage{xcolor}         \usepackage{amsmath}
\usepackage{amsthm}
\usepackage{mathtools}

\usepackage[capitalize]{cleveref}
\usepackage{caption}

\usepackage{pgfplots}
\usepackage{subcaption}
\usepackage{graphicx}
\usepackage{multirow}
\usepackage{array}
\usepackage{amssymb}

\usepackage{enumitem}
\usepackage{csquotes}

\def\vt{\mathbf}
\def\bs{\boldsymbol}
\def\KL{\mathrm{D}_{\mathrm{KL}}}
\def\Dn{\mathcal D_{\mathrm{x}}}
\def\En{\mathcal D_{\mathrm{y}}}
\def\Es{\mathcal D_{\mathrm{s}}}

\def\Dm{\mathcal E}
\def\KNIFE{\textsc{Knife}}
\def\SMILE{SMILE}
\def\TUBA{TUBA}
\def\NWJ{NWJ}
\def\Schraudolph{\textsc{Schrau.}}
\def\doe{\textsc{DoE}}
\def\adv{\text{Adv}}
\def\PPP{\mathcal P}
\def\XXX{\mathcal X}
\def\EEE{\mathcal E}
\def\RR{\mathbb R}

\def\defas{:=}
\def\ie{i.e.}

\def\dd{\mathrm{d}}

\def\eps{\varepsilon}
\newcommand{\norm}[1]{\Vert#1\Vert}
\newcommand{\Exp}[1]{\mathbb E[#1]}
\newcommand{\Prob}[1]{\mathrm P\{#1\}}
\newcommand{\htp}[2][]{\hat p_{#1}(#2)}

\newtheorem{theorem}{Theorem}
\newtheorem{lemma}[theorem]{Lemma}

\crefname{equation}{Equation}{Equations}
\Crefname{equation}{Equation}{Equations}
\crefformat{equation}{#2(#1)#3}
\crefrangeformat{equation}{#3(#1)#4--#5(#2)#6}
\crefmultiformat{equation}{#2(#1)#3}{ and~#2(#1)#3}{, #2(#1)#3}{ and~#2(#1)#3}
\Crefformat{equation}{#2(#1)#3}
\Crefrangeformat{equation}{#3(#1)#4--#5(#2)#6}
\Crefmultiformat{equation}{#2(#1)#3}{ and~#2(#1)#3}{, #2(#1)#3}{ and~#2(#1)#3}
\crefname{figure}{Figure}{Figures}
\Crefname{figure}{Figure}{Figures}

\definecolor{lightgray}{gray}{0.55}
\newcommand{\result}[2]{ #1 \color{lightgray}{\scriptsize{$\pm{#2}$}}}

\DeclareMathOperator{\nat}{nat}
\DeclareMathOperator{\vol}{vol}
\DeclareMathOperator{\MI}{I}
\DeclareMathOperator{\Ent}{h}

\DeclareMathOperator{\CE}{CE}
\DeclareMathOperator{\kernel}{\kappa}

\pgfplotsset{compat=1.17}

\begin{document}
\twocolumn[
\icmltitle{
A Differential Entropy Estimator for Training Neural Networks
}

\icmlsetsymbol{equal}{*}
\begin{icmlauthorlist}
\icmlauthor{Georg Pichler}{equal,wu}
\icmlauthor{Pierre Colombo}{equal,cs}
\icmlauthor{Malik Boudiaf}{equal,ets}
\icmlauthor{G\"unther Koliander}{w}
\icmlauthor{Pablo Piantanida}{pablo}
\end{icmlauthorlist}

\icmlaffiliation{wu}{Institute of Telecommunications, TU Wien, 1040 Vienna, Austria}
\icmlaffiliation{ets}{\'ETS Montreal, Quebec H3C 1K3, Canada}
\icmlaffiliation{cs}{Laboratoire des Signaux et Systèmes (L2S),Paris-Saclay CNRS CentraleSupélec, 91190 Gif-sur-Yvette, France}
\icmlaffiliation{pablo}{International Laboratory on Learning Systems (ILLS),  
Université McGill - ETS - MILA - CNRS - Université Paris-Saclay - CentraleSupélec , Montreal, Quebec, Canada}
\icmlaffiliation{w}{Acoustics Research Institute, Austrian Academy of Sciences, 1040, Vienna, Austria}
\icmlcorrespondingauthor{Georg Pichler}{\href{mailto:georg.pichler@ieee.org}{georg.pichler@ieee.org}}

\icmlkeywords{Machine Learning, ICML}

\vskip 0.3in
]

\printAffiliationsAndNotice{\icmlEqualContribution}

\begin{abstract}
  
  Mutual Information (MI) has been widely used as a loss regularizer for training neural networks. This has been particularly effective when learn disentangled or compressed representations of high dimensional data. However, differential entropy (DE), another fundamental measure of information,
    has not found widespread use in neural network training. Although DE offers a potentially wider range of applications than MI, off-the-shelf DE estimators are either non differentiable, computationally intractable or fail to adapt to changes in the underlying distribution. These drawbacks prevent them from being used as regularizers in neural networks training. To address shortcomings in previously proposed estimators for DE, here we introduce {\KNIFE}, a fully parameterized, differentiable kernel-based estimator of DE. The flexibility of our approach also allows us to construct {\KNIFE}-based estimators for conditional (on either discrete or continuous variables) DE, as well as MI. We empirically validate our method on high-dimensional synthetic data and further apply it to guide the training of neural networks for real-world tasks. Our experiments on a large variety of tasks, including visual domain adaptation, textual fair classification, and textual fine-tuning demonstrate the effectiveness of {\KNIFE}-based estimation. Code can be found at \url{https://github.com/g-pichler/knife}.

      \end{abstract}
\section{Introduction}

Learning tasks requires information~\citep{principe2006information} in the form of training data. Thus, information measures~\citep{Shannon1948Mathematical} (\textit{e.g.}\ entropy, conditional entropy and mutual information) have been a source of inspiration for the design of learning objectives in modern machine learning (ML) models~\citep{Linsker1989How,Torkkola2006}. Over the years, a plethora of estimators have been introduced to estimate the value of the aforementioned measures of information and they have been applied to many different problems, including  information and coding theory, limiting distributions, model selection, design of experiment and optimal prior distribution, data disclosure, and relative importance of predictors~\citep{https://doi.org/10.1111/j.1751-5823.2010.00105.x}. In these applications, traditional research focused on both developing new estimators and obtaining provable guarantees on the asymptotic behavior of these estimators~\citep{NIPS2012_c8ba76c2,Verdu2019Empirical}.

However, when used for training deep neural networks, additional \emph{requirements} need to be satisfied. In particular, the estimator needs to be differentiable w.r.t.\ the data distribution \textbf{(R1)}, computationally tractable \textbf{(R2)}, and able to rapidly adapt to changes in the underlying distribution \textbf{(R3)}.
For instance, Mutual Information (MI), a fundamental measure of dependence between variables, only became a popular (standalone or regularizing) learning objective for DNNs once estimators satisfying the above requirements were proposed~\citep{pmlr-v97-poole19a, 10.5555/2981345.2981371}. Although MI is notoriously difficult to estimate in high dimensions~\citep{PhysRevE.69.066138, DBLP:journals/corr/abs-2002-02851, pmlr-v108-mcallester20a}, these estimators have demonstrated promising empirical results in unsupervised representation learning \citep{NIPS2010_42998cf3, NIPS1991_a8abb4bb,hjelm2018learning, Tschannen2020On}, discrete/invariant  representations~\citep{pmlr-v70-hu17b,Ji_2019_ICCV}, generative modeling \citep{infogan,info_vae}, textual disentangling~\citep{dt_info,colombo2021novel}, and applications of the Information Bottleneck (IB) method~\citep{mahabadi2021variational,devlin2018bert, DBLP:journals/corr/AlemiFD016} among others. Compared to MI, Differential Entropy (DE) has received less attention from the ML community while offering a potentially wider range of applications than MI.

In this paper, we focus on the problem of differentiable, flexible and efficient DE estimation satisfying simultaneously \textbf{(R1)},\textbf{(R2)},\textbf{(R3)} as this quantity naturally appears in many applications (\textit{e.g.}\ reinforcement learning \citep{pmlr-v97-shyam19a, provably_efficient_entropy_maximization, impact_entropy_policy_optimization,emi}, IB \citep{DBLP:journals/corr/AlemiFD016}, mode collapse \citep{mine}) and can be used to estimate MI. Traditional estimators of DE often violate at least one of the requirements \textbf{(R1)} -- \textbf{(R3)} listed above (\textit{e.g.}\ $k$-nearest neighbor based estimators violate \textbf{(R1)}). Consequently, the lack of appropriate DE estimators for arbitrary data distributions forces deep learning researchers to either restrict themselves to special cases where closed-form expressions for DE are available \citep{pmlr-v97-shyam19a} or use  MI as a proxy \citep{mine}. In this work, we introduce a \underline{K}ernelized \underline{N}eural d\underline{IF}ferential \underline{E}ntropy (\KNIFE) estimator, that satisfies the aforementioned requirements and addresses limitations of existing DE estimators \citep{Schraudolph2004Gradient,pmlr-v108-mcallester20a}. Stemming from recent theoretical insights \citep{pmlr-v108-mcallester20a} that justify the use of DE estimators as building blocks to better estimate MI, we further apply {\KNIFE} to MI estimation. In the context of deep neural networks with high dimensional data (\textit{e.g.}\ image, text), {\KNIFE} achieves competitive empirical results in applications where DE or MI is required.

\subsection{Contributions}
Our work advances the field of differentiable, flexible and efficient DE estimation satisfying simultaneously \textbf{(R1)},\textbf{(R2)},\textbf{(R3)}.
\begin{enumerate}[wide, labelindent=0pt] \item We showcase limitation of the existing DE estimators proposed in \citep{Schraudolph2004Gradient,pmlr-v108-mcallester20a} with respect to desirable properties required for training deep neural networks. To address these shortcomings, we introduce {\KNIFE}, a fully learnable kernel-based estimator of DE. The flexibility of {\KNIFE} allows us to construct {\KNIFE}-based estimators for conditional DE, conditioning on either a discrete or continuous random variable.\vspace{-0.1cm}
    
\item We prove learnability under natural conditions on the underlying probability distribution. By requiring a fixed Lipschitz condition and bounded support we are not only able to provide an asymptotic result, but also a confidence bound in the case of a finite training set. This extends the consistency result by \citep{Ahmad1976Nonparametric}.\vspace{-0.1cm}

\item We validate on synthetic datasets (including multi-modal, non-Gaussian distributions), that {\KNIFE} addresses the identified limitations and outperforms existing methods on both DE and MI estimation. In particular, {\KNIFE} more rapidly adapts to changes in the underlying data distribution.\vspace{-0.1cm}

\item  We conduct extensive experiments on natural datasets (including text and images) to compare {\KNIFE}-based MI estimators to most recent MI estimators. First, we apply {\KNIFE} in the IB principle to fine-tune a pretrained language model. Using {\KNIFE}, we leverage a closed-form expression of a part of the training objective and achieve the best scores among competing MI estimators.
  Second, on fair textual classification, the {\KNIFE}-based MI estimator achieves near perfect disentanglement (with respect to the private, discrete label) at virtually no degradation of accuracy in the main task. Lastly, in the challenging scenario of visual domain adaptation, where both variables are continuous, {\KNIFE}-based MI estimation also achieves superior results. Code for {\KNIFE} and to reproduce results can be found at \url{https://github.com/g-pichler/knife}. \vspace{-0.1cm}
\end{enumerate}

\subsection{Existent Methods and Related Works}

\paragraph{DE estimation.} Existing methods for estimating DE  fit into one of three categories~\citep{Beirlant1997Nonparametric,hlavavckova2007causality,Verdu2019Empirical}: plug-in estimates~\citep{Ahmad1976Nonparametric,Gyoerfi1987Density}, estimates based on sample-spacings~\citep{Tarasenko1968evaluation}, and estimates based on nearest neighbor distances~\citep{Kozachenko1987Sample,Tsybakov1996Root,berrett2019efficient}. Our proposed estimator falls into the first category and we will thus focus here on previous work using that methodology.
Excellent summaries of all the available methods can be found in the works~\citep{Beirlant1997Nonparametric,hlavavckova2007causality,Wang2009Universal,Verdu2019Empirical}. In \citep{Ahmad1976Nonparametric}, a first nonparametric estimator of DE was suggested and theoretically analyzed.
It builds on the idea of kernel density estimation using Parzen-Rosenblatt windowing \citep{Rosenblatt1956Remarks,Parzen1962estimation}.
More detailed analysis followed \citep{Joe1989Estimation,Hall1993Estimation} but the estimator remained essentially unchanged.
Unfortunately, this classical literature is mostly concerned with appropriate regularity conditions that guarantee asymptotic properties of estimators, such as (asymptotic) unbiasedness and consistency.
Machine learning applications, however, usually deal with a fixed---often very limited---number of samples.

\paragraph{Differentiable DE estimation.} 
A first estimator that employed a differential learning rule was introduced in \citep{viola1996empirical}.
Indeed, the estimator proposed therein is optimized using stochastic optimization, it only used a single kernel with a low number of parameters.
An extension that uses a \emph{heteroscedastic} kernel density estimate, i.e., using different kernels at different positions, has been proposed in \citep{Schraudolph2004Gradient}.
Still the number of parameters was quite low and varying means in the kernels or variable weights were not considered.
Although the estimation of DE remained a topic of major interest as illustrated by recent works focusing on special classes of distributions \citep{kolchinsky2017estimating,chaubey2021estimation}  and nonparametric estimators \citep{ 6473887, NIPS2015_06138bc5,9496673},
the estimator introduced in \citep{Schraudolph2004Gradient}
was not further refined and hardly explored  in recent works.

\paragraph{Differentiable MI estimation.} In contrast, there has been a recent surge on new methods for the estimation of the closely related MI between two random variables.
The most prominent examples include  unnormalized energy-based variational lower bounds \citep{pmlr-v97-poole19a}, the lower bounds developed in \citep{convex_risk_minimization} using variational characterization of f-divergence, the MINE-estimator developed in \citep{mine} from the Donsker-Varadhan  representation of MI  which  can  be also  interpreted as an improvement of the plug-in estimator of~\citep{suzuki2008approximating},  the noise-contrastive based bound developed in \citep{mi_nce} and finally a contrastive upper bound~\citep{Cheng2020CLUB}. \citep{pmlr-v108-mcallester20a} point out shortcomings in other estimation strategies and introduce their own Differences of Entropies (\doe) method.

 \section{\KNIFE}

 In this section we identify limitations of existing entropy estimators introduced in \citep{Schraudolph2004Gradient,pmlr-v108-mcallester20a}. Subsequently, we present {\KNIFE}, which addresses these shortcomings. 
 
 \subsection{Limitations of Existing Differential Entropy Estimators}
 \label{sec:limit-exist-de}
Consider a continuous random vector $X \sim p$ in $\RR^d$. Our goal is to estimate the DE $\Ent(X) \coloneqq -\int p(x) \log p(x) \,\dd x$.
                     Given the intractability of this integral, we will rely on a Monte-Carlo estimate of $\Ent(X)$, using $N$ i.i.d.\ samples $\Dn = \{x_n\}_{n=1}^N$ to obtain
            \begin{equation} \label{eq:oracle}
                \widehat{\Ent}_{\textsc{Oracle}}(\Dn) \coloneqq - \frac{1}{N} \sum_{n=1}^N \log p(x_n).
            \end{equation}
            Unfortunately, assuming access to the true density $p$ is often unrealistic, and we will thus construct an estimate $\hat{p}$ that can then be plugged into \cref{eq:oracle} instead of $p$. If $\hat p$ is smooth, the resulting plug-in estimator of DE is differentiable \textbf{(R1)}.

            Assuming access to an additional---ideally independent---set of $M$ i.i.d.\ samples $\Dm = \{x_m'\}_{m=1}^M$, we build upon the Parzen-Rosenblatt estimator~\citep{Rosenblatt1956Remarks,Parzen1962estimation}
            \begin{equation}
              \label{eq:kernel0}
              \hat p(x; w, \Dm) = \frac{1}{w^d M} \sum_{m=1}^M \kappa\left(\frac{x - x_m'}{w} \right),
            \end{equation}
            where $w > 0$ denotes the bandwidth and $\kappa$ is a kernel density.
            The resulting entropy estimator when replacing $p$ in \cref{eq:oracle} by \cref{eq:kernel0} was analyzed in~\citep{Ahmad1976Nonparametric}.
            In \citep{Schraudolph2004Gradient}, this approach was extended using the kernel estimator
            \begin{equation}
              \label{eq:schraudolph}
              \hat p_{\Schraudolph}(x; \vt A, \Dm) \coloneqq \frac{1}{M} \sum_{m=1}^M \kappa_{A_m}(x - x_m') ,
            \end{equation}
            where $\vt A \coloneqq (A_1,\dots,A_M)$ are (distinct, diagonal) covariance matrices and $\kappa_A (x)= \mathcal N(x;0, A)$ is a centered Gaussian density with covariance matrix $A$.

            The {\doe} method of~\citep{pmlr-v108-mcallester20a} is a MI estimator that separately estimates a DE and a conditional DE. For DE, a simple Gaussian density estimate $\hat p_{\doe}(x; \bs\theta) = \kappa_A(x-\mu)$
            is used, where $\bs\theta = (A, \mu)$ are the training parameters, the diagonal covariance matrix $A$ and the mean $\mu$.
            
            While both {\Schraudolph} and {\doe} yield differentiable plug-in estimators for DE, they each have a major disadvantage. The strategy of \citep{Schraudolph2004Gradient} fixes the kernel mean values at $\EEE$, which implies that the method cannot adapt to a shifting input distribution \textbf{(R3)}. On the other hand, {\doe} allows for rapid adaptation, but its simple structure makes it inadequate for the DE estimation of multi-modal densities. We illustrate these limitations in \cref{sec:ent-estimation}.
            
             \subsection{{\KNIFE} Estimator}
 \label{sec:KNIFE}
 
            In {\KNIFE}, the kernel density estimate is given by
            \begin{equation}
              \label{eq:knife}
              \hat p_{\KNIFE}(x; \bs \theta) \coloneqq \sum_{m=1}^M u_m \kappa_{A_m}(x - b_m) ,
            \end{equation}
            where  $\bs \theta \coloneqq (\vt A, \vt b, \vt u)$ and the additional parameters $0 \le \vt u = (u_1, u_2, \dots, u_M)$ with $\vt 1 \cdot \vt u = 1$ and $\vt b = (b_1,\dots,b_M)$ are introduced.
            The covariance matrices $A_m \in \mathbb R^{d \times d}$, $m=1,\dots,M$ are symmetric and positive definite, but not necessarily diagonal.
            Note that the Gaussian densities $\kappa_{A_m}(x - b_m) = \mathcal N(x;b_m, A_m)$ are smooth functions. Thus, $\hat p_{\KNIFE}(x; \bs \theta)$ is a smooth function of $\bs \theta$, and so is our proposed plug-in estimator
            \begin{equation}
              \label{eq:hat-h}
              \widehat \Ent_{\KNIFE}(\Dn; \bs\theta) \coloneqq -\frac{1}{N} \sum_{n=1}^N \log \hat p_{\KNIFE}(x_n; \bs \theta).
            \end{equation}
            {\KNIFE} combines the ideas of \citep{Schraudolph2004Gradient,pmlr-v108-mcallester20a}. It is differentiable and able to adapt to shifting input distributions, while capable of matching multi-modal distributions.
            Thus, as we will see in synthetic experiments, incorporating $u_m$ and shifts $b_m$ in the optimization enables the use of ${\KNIFE}$ in non-stationary settings, where the distribution of $X$ evolves over time.

        \paragraph{Learning step: } Stemming from the observation that, by the Law of Large Numbers (LLN),
        \begin{equation}
\begin{split} \label{eq:ce_bound}
            \widehat \Ent_{\KNIFE}(\Dn, \bs\theta)  & \stackrel{\text{LLN}}{\approx}
            -\mathbb{E}~\big[\log\hat p_{\KNIFE}(X; \bs \theta)\big] \\ 
            & = \Ent(X) + \KL(p \| \hat p_{\KNIFE}(\,\cdot\,;\bs \theta)) \\  & \ge \Ent(X), 
                    \end{split}
        \end{equation}
        we propose to learn the parameters $\bs\theta$ by minimizing $\widehat \Ent_{\KNIFE}$, where $\Dm$ may be used to initialize $\vt b$. Although not strictly equivalent due to the Monte-Carlo approximation, minimizing $\widehat \Ent_{\KNIFE}$ can be understood as minimizing the Kullback-Leibler (KL) divergence in~\cref{eq:ce_bound}, effectively minimizing the gap between $\widehat \Ent_{\KNIFE}$ and $\Ent(X)$. In fact, $\widehat \Ent_{\KNIFE}$ can also be interpreted as the standard maximum likelihood objective, widely used in modern machine learning.  It is worth to mention that the {\KNIFE} estimator is fully differentiable with respect to $\bs \theta$ and the optimization can be tackled by any gradient-based method (e.g., Adam~\citep{Kingma2014Adam} or AdamW~\citep{loshchilov2017decoupled}).
        \subsection{Convergence Analysis}
        \label{sec:convergence-analysis}
        Note that the classical Parzen-Rosenblatt estimator $\widehat\Ent(\Dn; w)$, where~\cref{eq:kernel0} is plugged into~\cref{eq:oracle}, is a special case of \KNIFE. Thus, the convergence analysis provided in \citep[Thm.~1]{Ahmad1976Nonparametric} also applies and yields sufficient conditions for $\widehat \Ent_{\KNIFE}(\Dn, \bs\theta) \to \Ent(X)$.
        In \cref{sec:error-bound}, we extend this result and, assuming that the underlying distribution $p$ is compactly supported on $\XXX = [0,1]^d$ and $L$-Lipschitz continuous, the following \lcnamecref{thm:convergence} is proved.
        \begin{theorem}
          \label{thm:convergence}
          For any $\delta > 0$, there exists a function $\eps(N, M, w)$ such that, with probability at least $1-\delta$,
          $\big\lvert \widehat \Ent(\Dn; w) - \Ent(X)\big\rvert \le \eps(N, M, w)$.
          Additionally, $\eps(N, M, w) \to 0$ as ${M,}N \to \infty$ and $w\to 0$ if
          \begin{align}
            Nw &\to 0  &\text{and}&& 
                                    \frac{N^2 \log N}{w^{2d} M} &\to 0 ,
          \end{align}
          where $M$ and $N$ denote the number of samples in $\EEE$ and $\Dn$, respectively.
        \end{theorem}
        The precise assumptions for \cref{thm:convergence} and an explicit formula for $\eps(N, M, w)$ are given in \cref{thm:confidence_bound} in \cref{sec:error-bound}. For instance, \cref{thm:convergence} provides a bound on the speed of convergence for the consistency analysis in \citep[Thm.~1]{Ahmad1976Nonparametric}.
        Note, however, that this convergence result applies to the Parzen-Rosenblatt estimator and not to {\KNIFE}.
        
        \subsection{Estimating Conditional Differential Entropy and Mutual Information}
        \label{sec:cond_mi}
        Similar to \citep{pmlr-v108-mcallester20a}, the proposed DE estimator can be used to estimate other information measures. In particular, we can use {\KNIFE} to construct estimators of conditional DE and MI.
        When estimating the conditional DE and MI for a pair of random variables $(X,Y) \sim p$, we not only use $\Dn = \{x_n\}_{n=1}^N$, but also the according i.i.d.\ samples $\En = \{y_n\}_{n=1}^N$, where $(x_n, y_n)$ are drawn according to $p$.
        
        \paragraph{Conditional Differential Entropy.}
        We estimate conditional DE $\Ent(X|Y)$ by considering $\bs\theta$ to be a parameterized function $\bs\Theta(y)$ of $y$. Then all relations previously established naturally generalize and
        \begin{align} 
          \hat p_{\KNIFE}(x|y; \bs\Theta) & \coloneqq \hat p_{\KNIFE}(x; \bs\Theta(y)), \\
          \widehat \Ent_{\KNIFE}(\Dn|\En; \bs\Theta)&  \coloneqq  \frac{1}{N} \sum_{n=1}^N \log \frac{1}{\hat p_{\KNIFE}(x_n|y_n; \bs\Theta)}.
          \label{eq:hat_h_xgivy}
        \end{align}
        Naturally, minimization of~\cref{eq:hat_h_xgivy} is now performed over the parameters of $\bs\Theta$.
        If $Y$ is a continuous random variable, we use an artificial neural network $\bs\Theta(y)$, taking $y$ as its input.
          On the other hand,
          if $Y \in \mathcal{Y}$ is a discrete random variable, we have one parameter $\bs\theta$ for each $y \in \mathcal Y$, i.e., $\bs\Theta = \{\bs \theta_y\}_{y \in \mathcal Y}$ and
          $\hat p_{\KNIFE}(x|y; \bs\Theta) = \hat p_{\KNIFE}(x; \bs\Theta(y)) = \hat p_{\KNIFE}(x; \bs\theta_y)$.
        \paragraph{Mutual Information.}
        To estimate the MI between random variables $X$ and $Y$ (either discrete or continuous), recall that MI can be written as $\MI(X;Y) = \Ent(X) - \Ent(X|Y)$. Therefore, we use the marginal and conditional DE estimators \cref{eq:hat-h,eq:hat_h_xgivy} to build a {\KNIFE}-based MI estimator 
        \begin{align}
          \widehat \MI_{\KNIFE}(\Dn, \En; \bs\theta, \bs\Theta) &\coloneqq  \widehat{\Ent}_{\KNIFE}(\Dn; \bs{\theta}) \nonumber\\*
                                                                &\quad - \widehat \Ent_{\KNIFE}(\Dn|\En; \bs{\Theta}). \label{eq:MI}
        \end{align}

\section{Experiments using Synthetic Data}

\subsection{Differential Entropy Estimation}
\label{sec:ent-estimation}
In this \lcnamecref{sec:ent-estimation} we apply {\KNIFE} for DE estimation, comparing it to~\cref{eq:schraudolph}, the method introduced in~\citep{Schraudolph2004Gradient}, subsequently labeled ``\Schraudolph''. It is worth to mention that we did not perform the Expectation Maximization algorithm, as suggested in~\citep{Schraudolph2004Gradient}, but instead opted to use the same optimization technique as for {\KNIFE} to facilitate a fair comparison. 
\subsubsection{Gaussian Distribution}
\label{sec:gauss}
As a sanity check, we test {\KNIFE} on multivariate normal data in moderately high dimensions, comparing it to {\Schraudolph} and {\doe}, which we trained with the exact same parameters.
We estimate the entropy $\Ent(X) = \frac{d}{2} \log 2\pi e$ of $X \sim \mathcal N(\vt 0, I_d)$ for $d=10$ and $d=64$. The mean error and its empirical standard deviation are reported in \cref{tab:gaussian_results} over $20$ runs.
{\KNIFE} yielded the lowest bias and variance in both cases, despite {\doe} being perfectly adapted to matching a multivariate Gaussian distribution. Additional experimental details can be found in \cref{sec:appendix:gaussian-data}.

\begin{table}
  \begin{tabular}{rcc}
    \toprule
    $|h - \widehat h|$ & $d=10$ & $d=64$ \\
    \midrule
    {\doe} & $0.8388 \pm 1.0045$ & $3.3170 \pm 1.8281$\\
    {\Schraudolph} & $0.7301 \pm 0.0428$ & $9.8919 \pm 0.1604$ \\
    {\KNIFE} & $\mathbf{0.0461 \pm 0.0139}$ & $\mathbf{2.8045 \pm 0.0796}$ \\
    \bottomrule
  \end{tabular}
  \caption{Results on normal data with dimension $d$.}
  \label{tab:gaussian_results}
\end{table}

\begin{figure*}
    \begin{minipage}[t]{0.32\linewidth}
        \includegraphics[width=\textwidth]{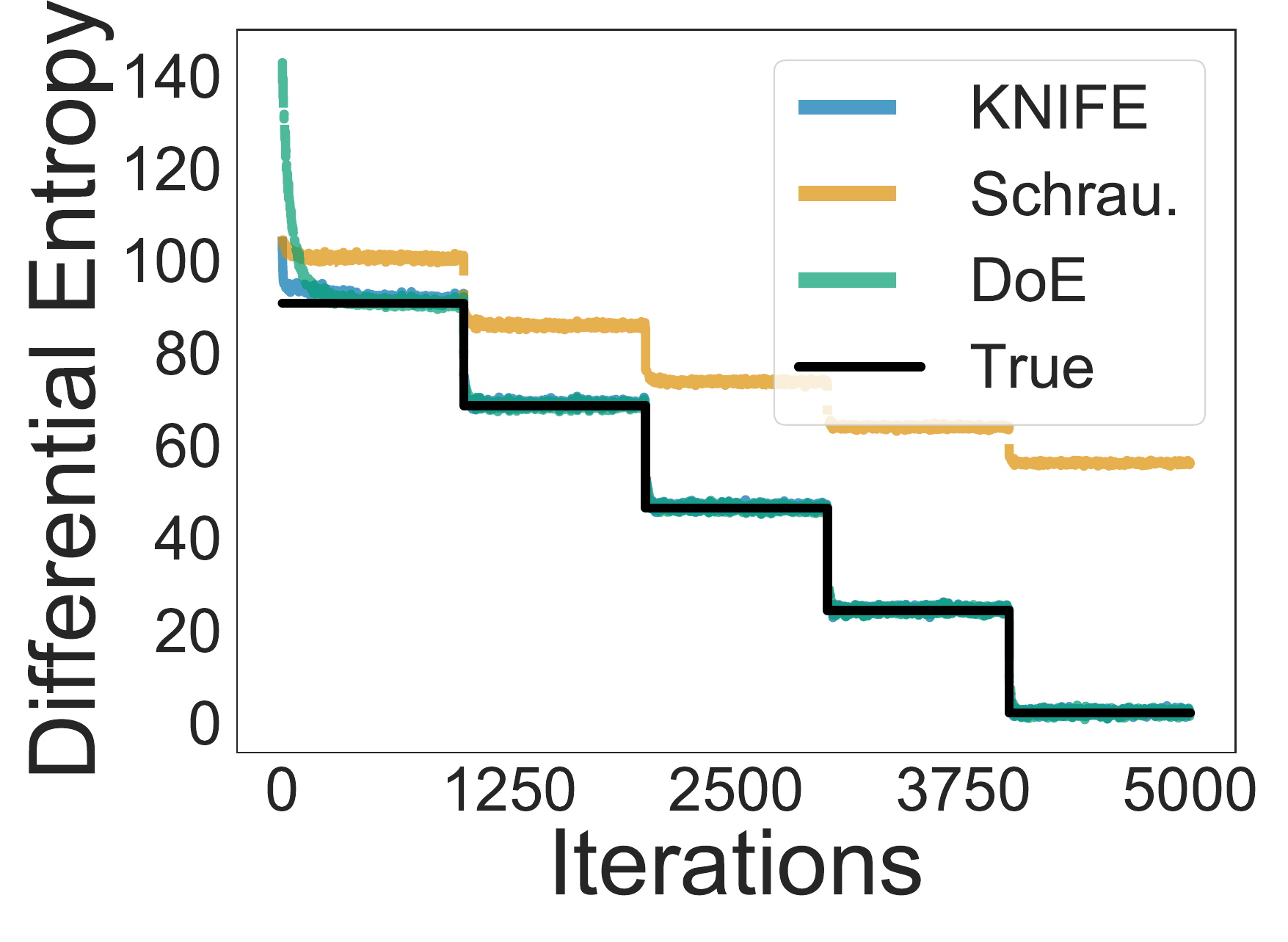}\vspace{-0.01cm}
        \caption{Estimating DE of Gaussian data with decreasing variance, i.e., $X_i \sim \mathcal N(\vt 0, a^{-i} I_d)$ for $d=64$ and $i = 0, \dots, 4$.}\label{fig:gauss}
    \end{minipage}\hfill\begin{minipage}[t]{0.65\linewidth}
    \centering
    \subcaptionbox*{}
  {\includegraphics[width=.4923\linewidth]{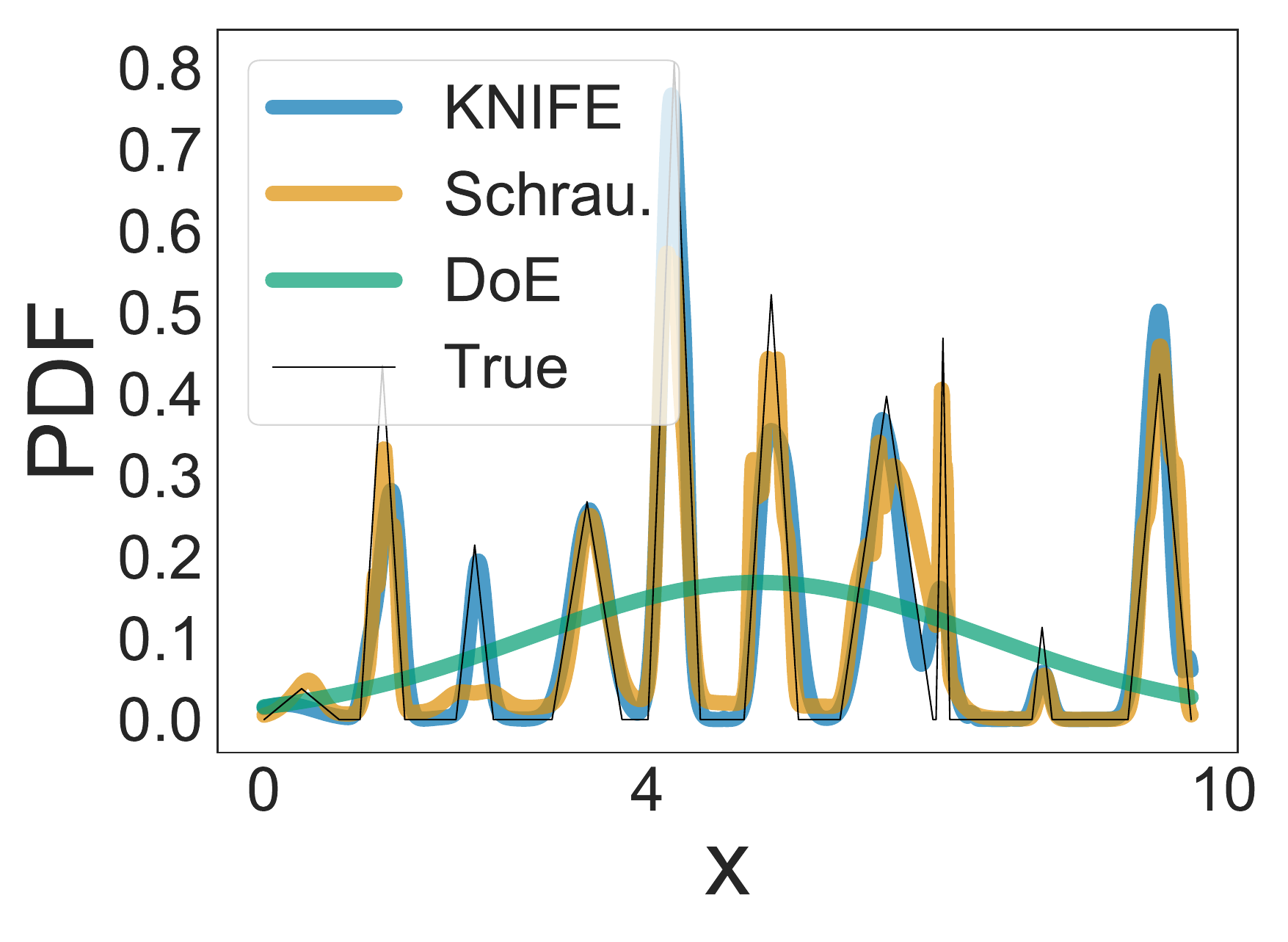}}
    \subcaptionbox*{}
  {\includegraphics[width=.4923\linewidth]{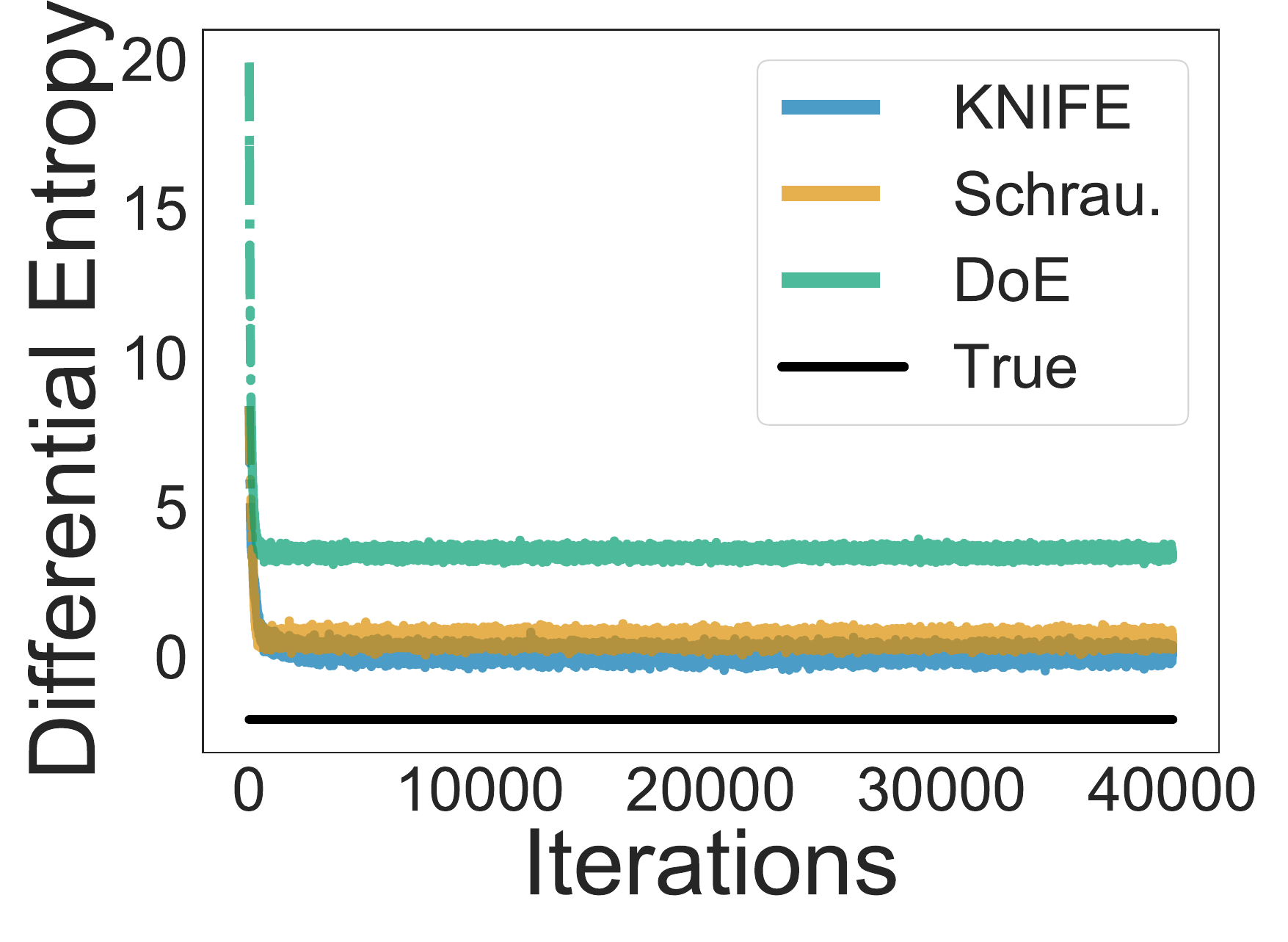}}\vspace{-0.8cm}
  \caption{\textbf{Left:} PDF when estimating DE of a triangle mixture in 1 dimension. \textbf{Right:} Training run when estimating DE of a $2$-component triangle mixture in 8 dimensions.}\label{fig:tirangle_mixture_expe}
\end{minipage}\vspace{-0.5cm}
\end{figure*}

In order to use a DE estimation primitive in a machine learning system, it must be able to adapt to a changing input distribution during training \textbf{(R3)}. 
As already pointed out in \cref{sec:limit-exist-de}, this is a severe limitation of {\Schraudolph}, as re-drawing the kernel support $\mathcal E$ can be either impractical or at the very least requires a complete re-training of the entropy estimator.
Whereas in \cref{eq:knife}, the kernel support $\vt b$ is trainable and it can thus 
adapt to a change of the input distribution. In order to showcase this ability, we utilize the approach of \citep{Cheng2020CLUB} and successively decrease the entropy, observing how the estimator adapts. We perform this experiment with data of dimension $d=64$ and  repeatedly multiply the covariance matrix of the training vectors with a factor of $a = \frac 12$.
The resulting entropy estimation is depicted in \cref{fig:gauss}. It is apparent that {\Schraudolph} suffers from a varying bias. The bias increases with decreasing variance, as the kernel support is fixed and cannot adapt as the variance of $\Dn$ shrinks.
{\doe} is perfectly adapted to a single Gaussian distribution and performs similar to \KNIFE. 
                                                                                               \subsubsection{Triangle Mixture}\label{sec:ent_triangle}
{\KNIFE} is able to cope with distributions that have multiple modes. While \cref{eq:schraudolph} is also capable of matching multi-modal distributions, {\doe} is unable to do so, as it approximates any distribution with a multivariate Gaussian.
We illustrate this by matching a mixture of randomly drawn triangle distributions.
The resulting estimated PDFs as well as the ground truth when estimating the entropy of a $1$-dimensional mixture of triangles with $10$ components can be observed in \cref{fig:tirangle_mixture_expe} (left).
With increasing dimension the difficulty of this estimation rises quickly as in $d$ dimensions, the resulting PDF of independent $c$-component triangle mixtures has $c^d$ modes.
To showcase the performance of {\KNIFE} in this challenging task, we 
ran $10$ training runs for DE estimation of $2$-component triangle mixtures in $8$ dimensions. An example training run is depicted in \cref{fig:tirangle_mixture_expe} (right).

\subsection{Mutual Information Estimation}
\label{sec:synthetic_mi_estimation}

\begin{figure*}
    \subcaptionbox*{}
  {
  \includegraphics[width=\textwidth]{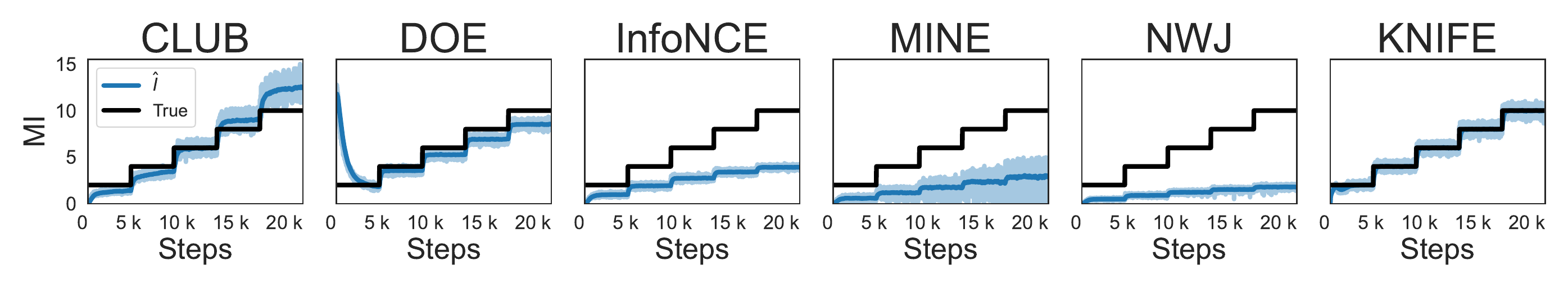}}\vspace{-.5cm}
  \subcaptionbox*{}
  {\includegraphics[width=\textwidth]{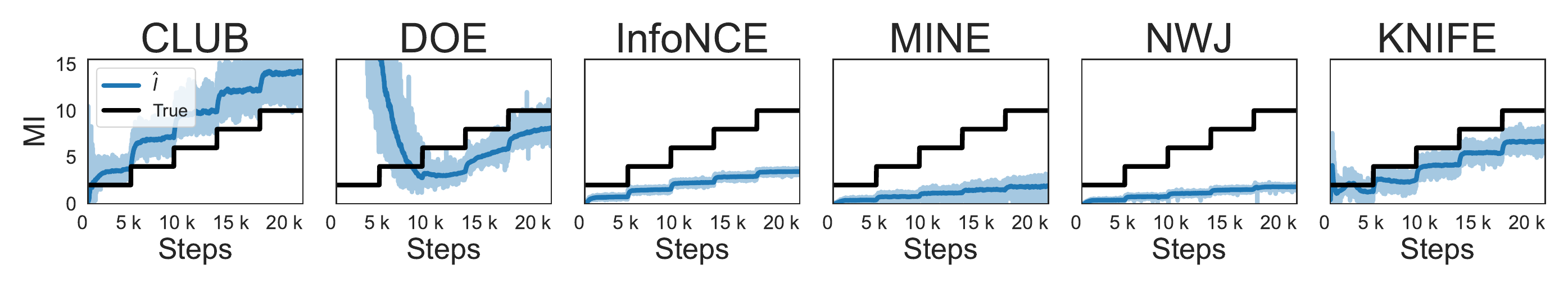}}\vspace{-.5cm}
    \subcaptionbox*{}
  {\includegraphics[width=\textwidth]{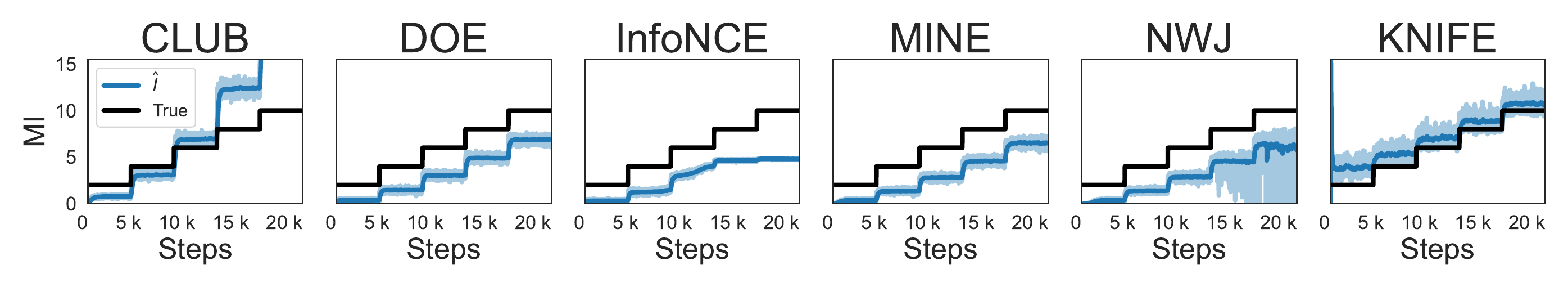}}\vspace{-.5cm}
  \caption{\textbf{Top:} Estimation of $\MI(X^d; Y^d)$, where $(X, Y)$ are multivariate Gaussian with correlation coefficient $\rho_i$ in the $i$-th epoch and $d=20$. \textbf{Middle:} Estimation of $\MI(X^d; (Y^3)^d)$. \textbf{Bottom:} Estimation of $\MI(X^d; Y^d)$ for uniform $(X, E)$ and $Y = \rho_i X + \sqrt{1-\rho_i^2}E$ in the $i$-th epoch. The best run out of 10 (by distance from the true MI at the end of training) is depicted. The dark curve shows an exponentially weighted moving average.}
  \label{fig:mi_gauss}
\end{figure*}

\paragraph{Multivariate Gauss.} We repeat the experiments in~\citep{Cheng2020CLUB}, stepping up the MI $\MI(X^d; Y^d)$ between $d$ i.i.d.\ copies of joint normal random variables $(X,Y)$ by increasing their correlation coefficient, i.e., $(X,Y)$ are multivariate Gaussian with correlation coefficient $\rho_i$ in the $i$-th epoch.
A training run is depicted in the top of \cref{fig:mi_gauss}.
As in~\citep{Cheng2020CLUB}, we also repeat the experiment, applying a cubic transformation to $Y$. The estimation of MI between $d$ i.i.d.\ copies of $X$ and $Y^3$ can be observed in the middle row of \cref{fig:mi_gauss}. The MI is unaffected by this bijective transformation.
In \cref{sec:app:MI_estimation}, the bias and variance are depicted separately.

\paragraph{Sum of Uniformly Distributed Variables}
\label{sec:uniform}
In order to test the ability of {\KNIFE} to adapt to distributions substantially different from the Gaussian kernel shape, we apply it in MI estimation of $\MI(X^d; Y^d)$ with uniformly distributed data. To this end, let $X$ and $E$ be centered, uniformly distributed random variables with $\mathbb E[X^2] = \mathbb E[E^2] = 1$ and define 
$Y = \rho_i X + \sqrt{1-\rho_i^2}E$  in the $i$-th epoch. One training run with $d=20$ is shown in \cref{fig:mi_gauss} (bottom). Details about the source distribution as well as details of the experiments can be found in \cref{sec:app:MI_estimation}.

\section{Experiments on Natural Data}\label{sec:natural_data}

In this section, we benchmark our proposed {\KNIFE}-based MI estimator on three practical applications, spanning textual and visual data. We reproduce and compare our method to the most recent MI estimators including MINE~\citep{mine},  NWJ~\citep{convex_risk_minimization}, InfoNCE~\citep{mi_nce}, CLUB~\citep{Cheng2020CLUB}, and {\doe}~\citep{pmlr-v108-mcallester20a}. We do not explicitly include the SMILE estimator~\citep{song2019understanding} in our comparison as it has the same gradient as NWJ.\footnote{Although not detailed in the original paper, {\SMILE}'s clipping is only used for estimation. For backpropagation, the gradients of the standard {\NWJ} bound are used. This is evident in the authors' original implementation at \href{https://github.com/ermongroup/smile-mi-estimator/blob/afa72e03f4c80a0997eb8f6b95d7f96bdf916ea7/estimators.py\#L123}{https://github.com/ermongroup/smile-mi-estimator/.../estimators.py\#L123}.}

\textbf{Common notation}: In all following applications, we will use $\Phi_{\bs{\psi}} \colon \mathcal{X} \rightarrow \mathcal{Z}$ to denote an encoder, where $\mathcal{X}$ is the raw input space (i.e., texts or images), and $\mathcal{Z}$ denotes a lower dimensional continuous feature space. 
Additionally, we will use $C_{\bs{\psi}} \colon \mathcal{Z} \rightarrow \mathcal{Y}$ to denote a shallow classifier from the latent space $\mathcal{Z}$ to a discrete or continuous target space $\mathcal{Y}$ for classification or regression, respectively. We will use $\bs\psi$ to denote the parameters of both models, $\Phi_{\bs{\psi}}$ and $C_{\bs\psi}$. CE denotes the cross entropy loss.

\subsection{Information Bottleneck for Language Model Finetuning}

IB has  recently been applied to fine-tune large-scale pretrained models \citep{mahabadi2021variational} such as BERT \citep{devlin2018bert} and aims at suppressing irrelevant features in order to reduce overfitting.
\paragraph{Problem statement.} Given a textual input $X \in \mathcal{X}$ and a target label $Y \in \mathcal{Y}$, the goal is to learn the encoder $\Phi_{\bs{\psi}}$ and  classifier $C_{\bs{\psi}}$, such that $\Phi_{\bs{\psi}}(X)$ retains little information about $X$, while still producing discriminative features, allowing the prediction of $Y$. Thus, the loss of interest is:
\begin{equation}\label{eq:bottleneck}
  \mathcal{L} =  \lambda \cdot \underbrace{\MI(\Phi_{\bs{\psi}} (X);X)}_{\mathclap{\textrm{compression term}}} \;\;-\;\;  \underbrace{\MI(\Phi_{\bs{\psi}}(X);Y)}_{\mathclap{\textrm{downstream  term}}} ,
\end{equation} where $\lambda$ controls the trade-off between the downstream and the compression terms.

\paragraph{Setup.} Following \citep{mahabadi2021variational} (relying on VUB), we work with the VIBERT model, which uses a Gaussian distribution as prior. 
$\Phi_{\bs{\psi}}$ is implemented as a stochastic encoder $\Phi_{\bs{\psi}} (X) = Z\sim \mathcal{N}(\mu_{\bs{\psi}}(X),\Sigma_{\bs{\psi}}(X))$. Details on the architecture of $\mu_{\bs{\psi}}$ and $\Sigma_{\bs{\psi}}$ can be found in \cref{sec:expdetailssupp}. 
The classifier $C_{\bs{\psi}}$ is composed of dense layers. To minimize $\mathcal L$, the second part of the objective \cref{eq:bottleneck} is bounded using the variational bound from \citep{10.5555/2981345.2981371}. Since we use a Gaussian prior, $\Ent(Z|X)$ can be expressed in closed form.\footnote{$\Ent(Z|X) =  \frac12\ln|\Sigma_{\bs{\psi}}(X)| + \frac{d}{2}\ln(2\pi e),$ where $d$ is the dimension of $X$ and $|\cdot|$ denotes the determinant.} 
Thus, when using {\KNIFE},  $\MI(X;Z)=\Ent(Z) - \Ent(Z|X)$ can be estimated by using $\widehat{\Ent}_{\KNIFE}$ to estimate $\Ent(Z)$.
We compare this {\KNIFE}-based MI estimator with  aforementioned MI estimators and the variational upper bound (VUB).
For completeness, we also compare against a BERT model trained by direct minimization of a CE loss.

We closely follow the protocol of \citep{mahabadi2021variational} and work on the GLUE benchmark \citep{wang2018glue} originally composed of 5 datasets. 
However, following \citep{mahabadi2021variational}, we choose to finetune neither on WNLI \citep{morgenstern2015winograd} nor on CoLA \citep{warstadt2019neural} due to reported flaws in these datasets.
The evaluation is carried out on the standard validation splits as the test splits are not available. 
Following standard practice \citep{liu2019roberta,yang2019xlnet}, we report
the accuracy and the F1 for MRPC, the accuracy for RTE and the Pearson and Spearman correlation coefficient for STS-B.

\paragraph{Results.}
\cref{tab:result_glue} reports our results on the GLUE benchmark. We observe that {\KNIFE} obtains the best results on all three datasets and the lowest variance on MRPC and STS-B. The use of a Gaussian prior in the stochastic encoder $\Phi_{\bs\psi}$ could explain the observed improvement of {\KNIFE}-based estimation over  MI-estimators such as CLUB, InfoNCE, MINE, \doe, or NWJ.

\begin{table}
  \centering
  \resizebox{0.45\textwidth}{!}{
    \begin{tabular}{cccccc}
      \toprule
      &  \multicolumn{2}{c}{MRPC}  & \multicolumn{2}{c}{STS-B}   & RTE\\
      \cmidrule(lr){2-3} \cmidrule(lr){4-5}  \cmidrule(lr){6-6}
      &       F1 &     Accuracy&       Pearson&Spearman     &  Accuracy \\
      \midrule
      BERT          & \result{83.4}{0.9}  &\result{88.2}{0.7}  & \result{89.2}{0.4}  & \result{88.8}{0.4} & \result{69.4}{0.4} \\
            CLUB                 & \result{85.0}{0.4}  & \result{89.0}{0.7}   & \result{89.7}{0.2}    & \result{89.4}{0.1}  & \result{70.7}{0.1}  \\
    InfoNCE       &\result{84.9}{0.8}   & \result{88.9}{0.6}  & \result{89.4}{0.4}   & \result{89.7}{0.6}  & \result{70.6}{0.1}  \\
      MINE         & \result{80.0}{2.5}  & \result{85.0}{0.9}  & \result{88.0}{0.7}   & \result{88.0}{0.6}  & \result{69.0}{0.9} \\
      NWJ        & \result{84.6}{0.8}   & \result{88.1}{0.7}   & \result{89.8}{0.1}   & \result{89.6}{0.2} & \result{69.6}{0.7}  \\
            VIBERT 
                      & \result{85.1}{0.5}  & \result{89.1}{0.3}  &\result{90.0}{0.2}   &\result{89.5}{0.3} & \result{70.9}{0.1} \\
                        {\doe}                         & \result{84.1}{0.2}  & \result{88.3}{0.2}  &\result{89.6}{0.2}   &\result{89.5}{0.2} & \result{69.6}{0.2} \\
      {\KNIFE}        & \result{\textbf{85.3}}{0.1}  & \result{\textbf{90.1}}{0.1}  & \result{\textbf{90.3}}{0.0}   & \result{\textbf{90.1}}{0.0}  & \result{\textbf{72.3}}{0.2}    \\
      \bottomrule
    \end{tabular}
  }            \caption{Fine-tuning on GLUE. Following \citep{lee2019mixout,dodge2020fine}, mean and variance are computed for 10 seeds. VIBERT is similar to VUB \citep{DBLP:journals/corr/AlemiFD016}.}\label{tab:result_glue}
\end{table}

\subsection{Fair Textual Classification}
In fair classification, we would like the model to take its decision without utilizing
private information such as gender, age,
or race.
For this task, MI can be minimized to disentangle the output of the encoder $Z$ and a private label $S \in \mathcal{S}$ (e.g., gender, age, or race).

\paragraph{Problem Statement.}
Given an input text $X$, a discrete target label $Y$ and a private label $S$, the loss is given by
\begin{equation}\label{eq:all_loss}
  \mathcal{L} =  \underbrace{\CE(Y; \Phi_{\bs{\psi}}(X)) }_{\textrm{downstream task}} +  \lambda \cdot \underbrace{\MI(\Phi_{\bs{\psi}}(X);S)}_{\textrm{disentangled}},
\end{equation}
where $\lambda$ controls the trade-off between minimizing MI and CE loss. In this framework, a classifier is said to be fair or to achieve perfect privacy if no statistical information about $S$ can be extracted from $\Phi_{\bs{\psi}}(X)$ by an adversarial classifier. 
Overall, a good model should achieve high accuracy on the main task (i.e.,  prediction of $Y$) while removing information about the protected attribute $S$. This information is measured by training an offline classifier to recover the protected attribute $S$ from $\Phi_{\bs{\psi}}(X)$.

\paragraph{Setup.}
We compute the second term of \cref{eq:all_loss} with competing MI estimators, as well as the model from \citep{elazar2018adversarial}, which will be referred to as ``{\adv}'',  as it utilizes an adversary to recover the private label from the latent representation $Z$. 
For \KNIFE-based MI estimation, we use two DE estimators (as $S$ is a binary label), following the approach outlined in \cref{sec:cond_mi}. All derivations are detailed in \cref{sec:expdetailssupp}. 
\newline We follow the experimental setting from \citep{elazar2018adversarial,barrett2019adversarial} and use two datasets from the DIAL corpus \citep{blodgett2016demographic} (over 50 million tweets) where the protected attribute $S$ is the race and the main labels are sentiment or mention labels. The mention label indicates whether a tweet is conversational or not. 
We follow the official split using $160\,000$ tweets for training and two additional sets composed of $10\,000$ tweets each for development and testing. In all cases, the labels $S$ and $Y$ are binary and balanced, thus a random guess corresponds to 50\%  accuracy.

\begin{figure*}
  \centering
    \subcaptionbox{$Y$ (sentiment) \label{fig:fair_classification_main_sentiment}}
  {\includegraphics[width=0.24\textwidth]{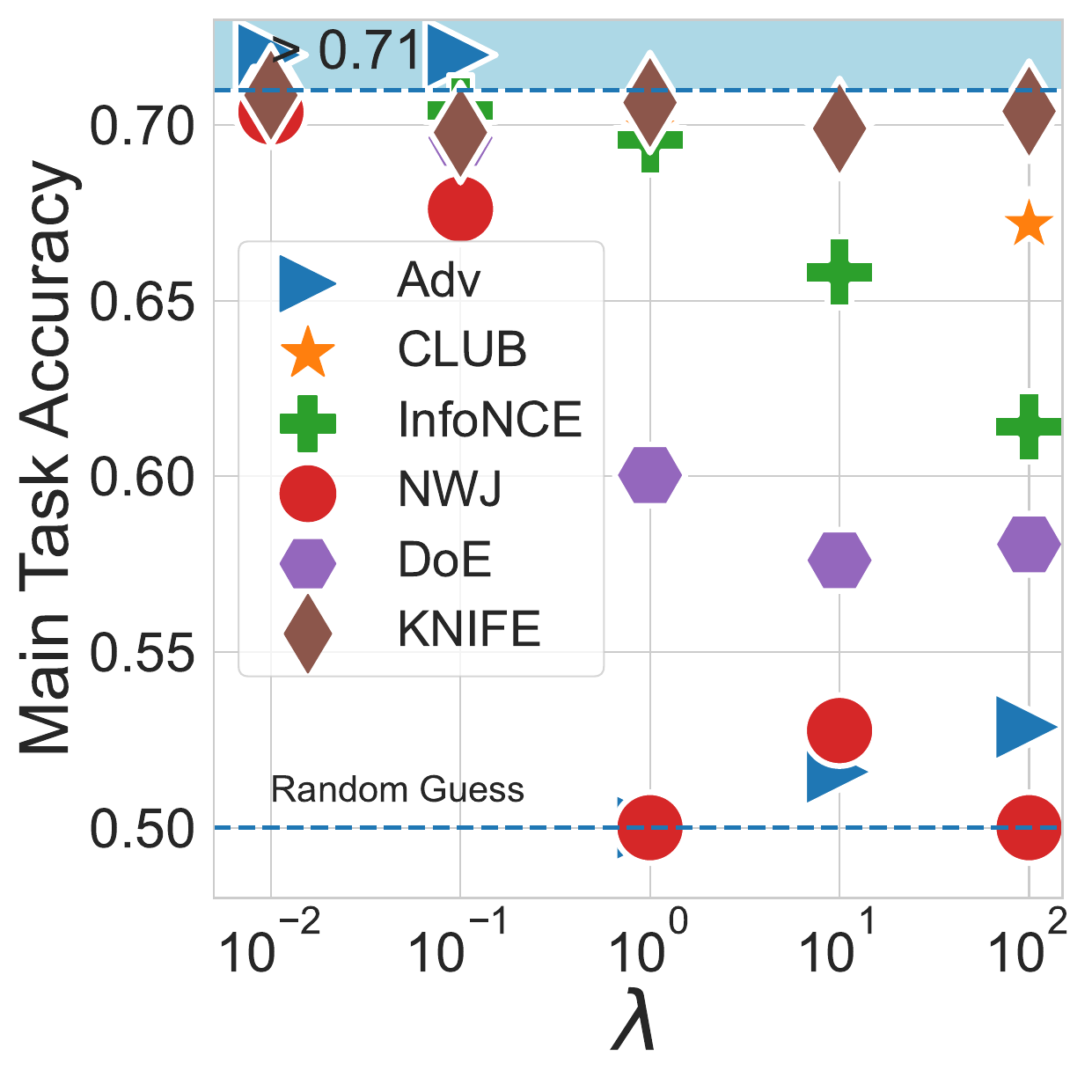}}\subcaptionbox{$S$ (sentiment) \label{fig:fair_classification_private_sentiment}}
  {\includegraphics[width=0.24\textwidth]{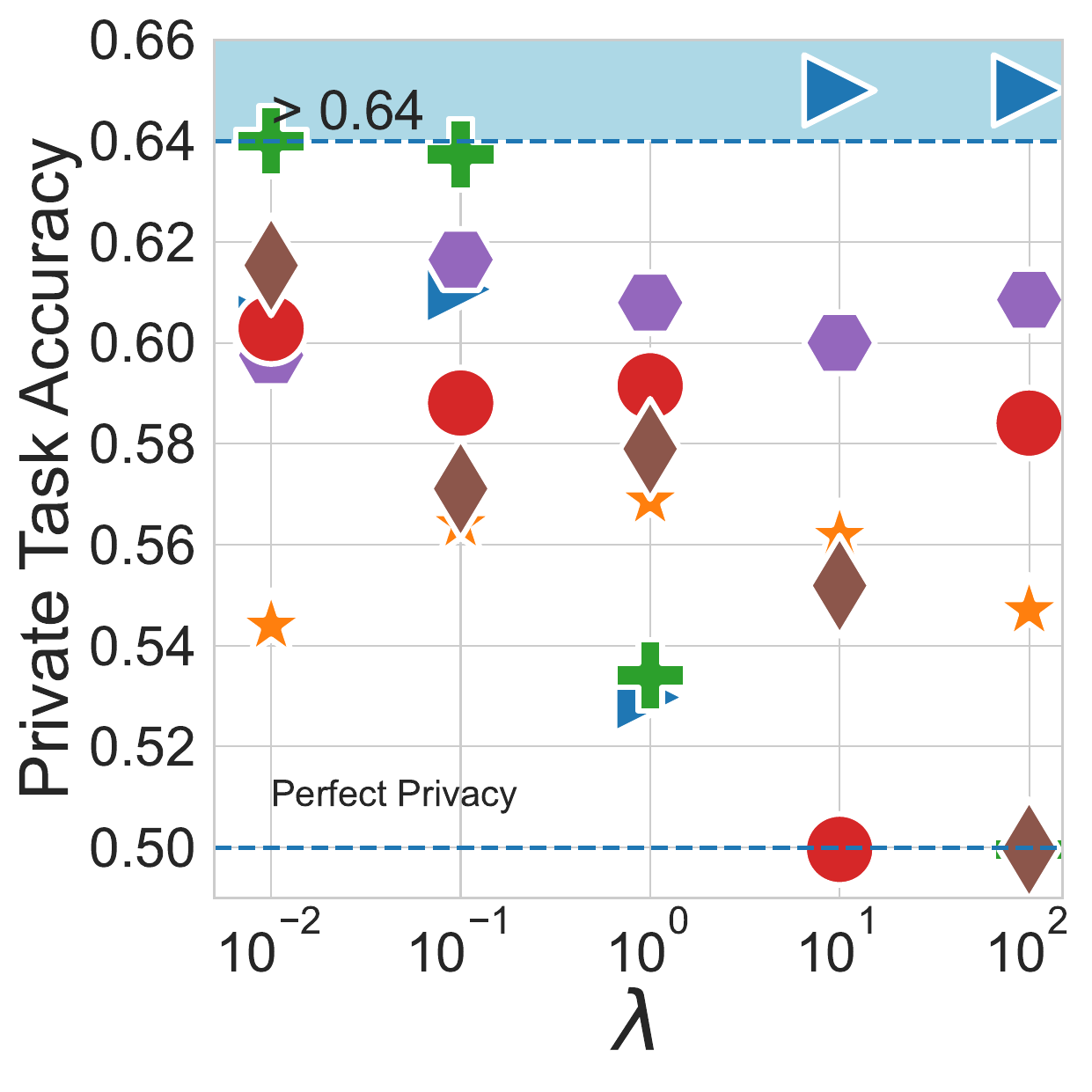}
  }\subcaptionbox{$Y$ (mention) \label{fig:fair_classification_main_mention}}
  {\includegraphics[width=0.24\textwidth]{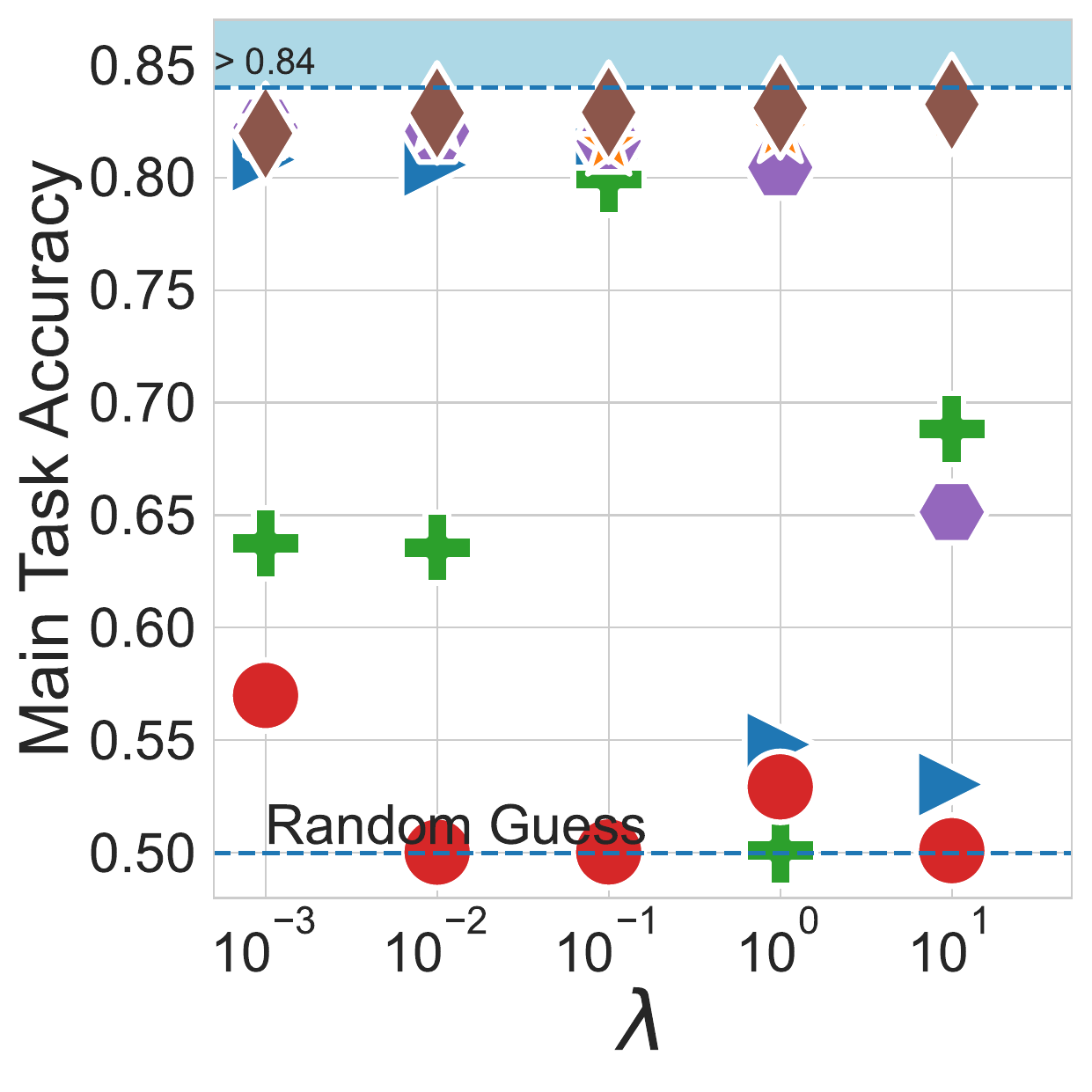}}
    \subcaptionbox{$S$ (mention)\label{fig:fair_classification_private_mention}}{ \includegraphics[width=0.24\textwidth]{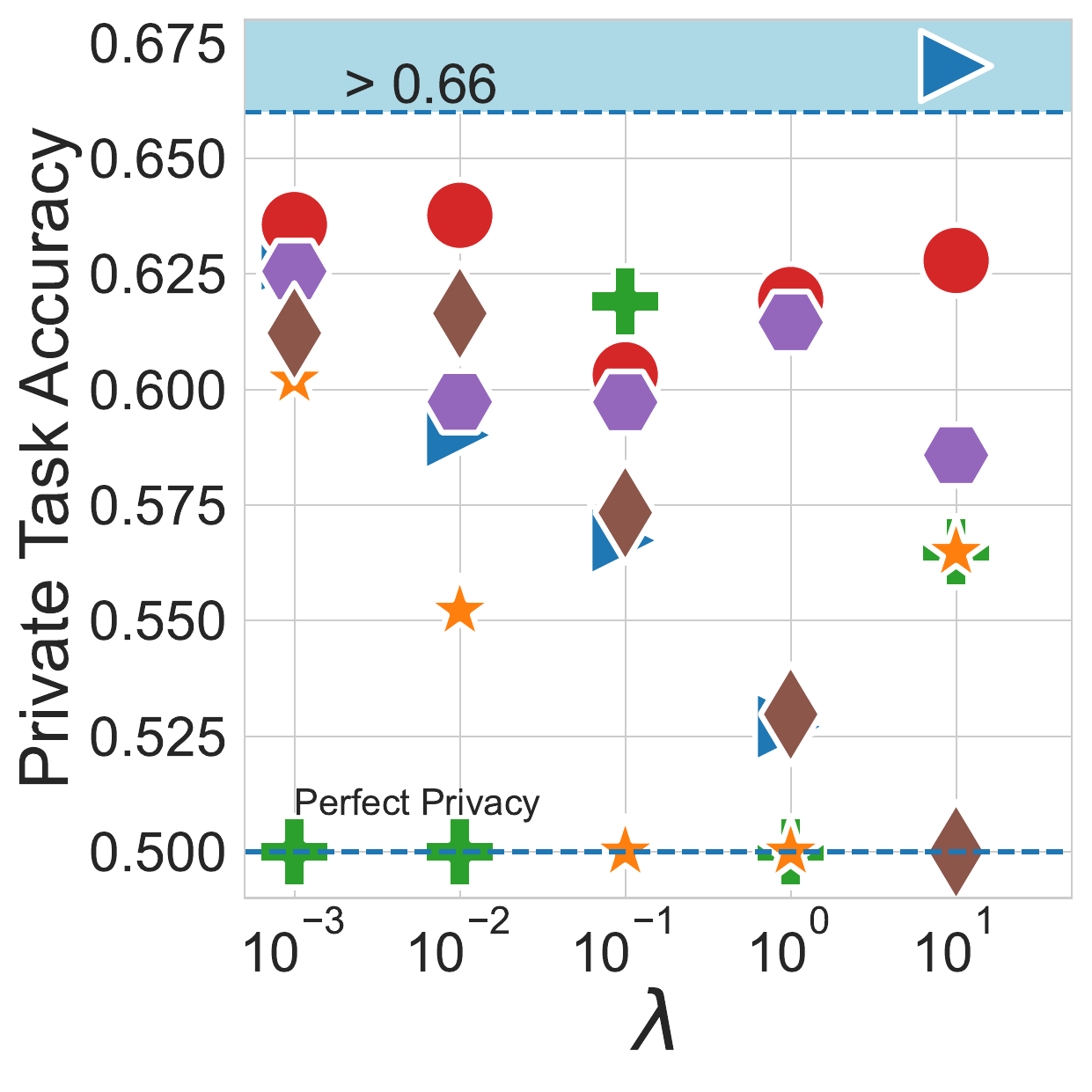}
  }
  \caption{Results on the fair classification task for both main (\cref{fig:fair_classification_main_sentiment,fig:fair_classification_main_mention}) and private task (\cref{fig:fair_classification_private_sentiment,fig:fair_classification_private_mention}) for both mention and sentiment labels. Results of MINE are not reported because of instabilities that prevent the network from converging.   \Cref{fig:fair_classification_private_sentiment,fig:fair_classification_private_mention} are obtained by training an offline classifier to recover the protected attribute $S$ from $\Phi_{\bs\psi}(X)$.}\label{fig:fair_classification_main}
\end{figure*}

\paragraph{Results.} Results on fair classification are displayed in~\cref{fig:fair_classification_main}. The upper dashed lines represent the (private and main) task accuracies when training a model with only the $\CE$ loss (case $\lambda = 0$ in \cref{eq:all_loss}). This shows that the learned encoding $\Phi_{\bs{\psi}}(X)$ contains information about the protected attribute when training is only performed for the main task. 
On both the sentiment and mention task, we observe that a \KNIFE-based estimator can achieve perfect privacy (\cref{fig:fair_classification_private_sentiment,fig:fair_classification_private_mention}) with nearly no accuracy loss in the main task (\cref{fig:fair_classification_main_sentiment,fig:fair_classification_main_mention}). 
The other MI estimators exhibit different behavior. For sentiment labels, most MI estimators fail to reach perfect privacy (CLUB, NWJ, \doe, and {\adv}) while others (InfoNCE) achieve perfect privacy while degrading the main task accuracy (10\% loss on main accuracy).
For mention labels, CLUB can also reach perfect privacy with almost no degradation in the main task. 
Overall, it is worth noting that {\KNIFE}-based MI estimation enables better control of the disentanglement than the reported baselines.

\subsection{Unsupervised Domain Adaptation}
\begin{table*}\centering
  \resizebox{0.7\textwidth}{!}{
    \begin{tabular}{cccccccc}
      \toprule
      & M $\rightarrow$ MM & S $\rightarrow$ C & U $\rightarrow$ M & M $\rightarrow$ U & C $\rightarrow$ S & SV $\rightarrow$ M &   Mean  \\
      \cmidrule{2-8}
      Source only       &       \result{51.9}{0.8} &   \result{58.3}{0.2} &     \result{91.1}{0.7} &        \result{93.5}{0.6} &        \result{\textbf{72.3}}{0.5} &     \result{54.7}{2.8} &   \result{70.3}{0.9} \\
      \midrule
      \textsc{CLUB}              &        \result{79.1}{2.2} & \result{59.9}{1.9} &     \result{96.0}{0.2} &          \result{96.8}{0.5} &        \result{71.6}{1.3} &     \result{83.8}{3.4} &  \result{81.2}{1.7} \\
      {\doe}                     &        \result{\textbf{82.2}}{2.6} & \result{58.9}{0.8} &     \result{97.2}{0.3} &          \result{94.2}{0.9} &        \result{68.8}{1.4} &     \result{86.4}{5.4} &  \result{81.3}{1.9} \\
      \textsc{InfoNCE}           &        \result{77.3}{0.5} &          \result{61.0}{0.1} &     \result{97.4}{0.2} &          \result{97.0}{0.3} &        \result{70.6}{0.8} &      \result{89.2}{4.1} &   \result{82.1}{1.0} \\
      \textsc{MINE}              &        \result{76.7}{0.4} &          \result{61.2}{0.3} &      \result{\textbf{97.7}}{0.1} &       \result{97.3}{0.1} &        \result{70.8}{1.0} &     \result{91.8}{0.8} &  \result{82.6}{0.4} \\
      \textsc{NWJ}               &        \result{77.1}{0.6} &          \result{61.2}{0.3} &      \result{97.6}{0.1} &           \result{97.3}{0.5} &        \result{72.1}{0.7} &     \result{91.4}{0.8} &    \result{82.8}{0.5} \\
      {\KNIFE}                   &        \result{78.7}{0.7} &          \result{\textbf{61.8}}{0.5} &  \result{\textbf{97.7}}{0.3} &   \result{\textbf{97.4}}{0.4} &   \result{71.2}{1.8} &     \result{\textbf{93.2}}{0.2} &  \result{\textbf{83.4}}{0.6} \\
      \bottomrule
    \end{tabular}
  }
    \caption{Domain adaptation results: M (MNIST), MM (MNIST M), U (USPS), SV (SVHN), C (CIFAR10) and S (STL10). Results are averaged over 3 seeds.}
  \label{tab:da_results}
\end{table*}
In unsupervised domain adaptation, the goal is to transfer knowledge from the source domain ($S$) with a potentially large number of labeled examples to a target domain ($T$), where only unlabeled examples are available.

\paragraph{Problem Statement.}
The learner is given access to labeled images from a source domain  $(x_s, y) \sim (X_S,Y) \in \mathcal{X}_{S} \times \mathcal{Y}$ and unlabeled images from a target domain $x_t \sim X_T \in \mathcal{X}_{T}$. The goal is to learn a classification model $\{\Phi_{\bs{\psi}}, C_{\bs{\psi}}\}$ that generalizes well to the target domain. 
Training models on the supervised source data only results in domain-specific latent representations $\Phi_{\bs{\psi}}(X)$ leading to poor generalization (when $X$ is chosen randomly from $\{X_S, X_T\}$).
In order to make the latent representations as domain-agnostic as possible, we follow the information-theoretic method proposed by \citep{gholami2020unsupervised}, and used in \citep{Cheng2020CLUB}. The idea is to learn an additional binary model $\{\Phi^d_{\bs{\nu}}, C^d_{\bs{\nu}}\}$, whose goal it is to guess the domain $D \in \{0, 1\}$ of $X$. 
The latent representation learned by $\Phi^d_{\bs{\nu}}$ will therefore contain all the domain-specific information that we would like the main encoder $\Phi_{\bs{\psi}}$ to discard. 
In other words, we would like $\Phi_{\bs{\psi}}(X)$ and $\Phi^d_{\bs{\nu}}(X)$ to be completely disentangled, which naturally corresponds to the minimization of $\MI(\Phi_{\bs{\psi}}(X);\Phi^d_{\bs{\nu}}(X))$.
Concretely, the domain classifier is trained to minimize the $\CE$ between domain labels $D$ and its own predictions, whereas the main classifier is trained to properly classify support samples while minimizing the MI between $\Phi_{\bs{\psi}}(X)$ and $\Phi^d_{\bs{\nu}}(X)$. Using $f_{\bs{\nu}}^d \coloneqq C_{\bs{\nu}}^d \circ \Phi^d_{\bs{\nu}}$ and $f_{\bs{\psi}}\coloneqq C_{\bs{\psi}} \circ \Phi_{\bs{\psi}}$, the objectives are
\vspace*{-.3\baselineskip}
\begin{align}
  \label{eq:domaine}
  \min_{{\bs{\nu}}} & ~ \CE(D; f_{\bs{\nu}}^d(X))  \; \text{and}  \\
  \min_{{\bs{\psi}}} & ~ \CE(Y; f_{{\bs{\psi}}}(X_S)) + \lambda \cdot \MI(\Phi_{\bs{\psi}}(X);\Phi_{\bs{\nu}}^d(X)).
\end{align}

\paragraph{Setup.} The different MI estimators are compared based on their ability to guide training by estimating $\MI(\Phi_{\bs{\psi}}(X);\Phi^d_{\bs{\nu}}(X))$ in \cref{eq:domaine}.  We follow the setup of \citep{Cheng2020CLUB} as closely as possible, and consider a total of 6 source/target scenarios formed with MNIST \citep{mnist}, MNIST-M \citep{mnist_m}, SVHN \citep{svhn}, CIFAR-10 \citep{cifar10}, and STL-10 \citep{stl10} datasets. We reproduce all methods and allocate the same budget for hyper-parameter tuning to every method. The exhaustive list of hyper-parameters can be found in \cref{sec:expdetailssupp}.

\paragraph{Results.} Results are presented in \cref{tab:da_results}. The \KNIFE-based estimator is able to outperform MI estimators in this challenging scenario where both $\Phi_{\bs{\psi}}(X)$ and $\Phi^d_{\bs{\nu}}(X)$ are continuous.

\section{Concluding Remarks}

We introduced {\KNIFE}, a fully learnable, differentiable kernel-based estimator of differential entropy, designed for deep learning applications. We constructed a mutual information estimator based on {\KNIFE} and showcased several applications. {\KNIFE} is a general purpose estimator and does not require any special properties of the learning problem. It can thus be incorporated as part of any training objective, where differential entropy or mutual information estimation is desired. In the case of mutual information, one random variable may even be discrete. 

Despite the fundamental challenges in the problem of differential entropy estimation, beyond limitations arising from the use of a finite number of samples, {\KNIFE} has demonstrated promising empirical results in various representation learning tasks.

Future work will focus on improving the confidence bounds given in \cref{thm:convergence}. In particular, tailoring them towards {\KNIFE}  using tools from \citep{10.1214/aos/1176324452, NIPS2014_af5afd7f}. Another potential extension is direct estimation of the gradient of entropy, when $\hat p_{\KNIFE}(x; \bs \theta)$ has been learned \citep{mohamed2020monte,song2020sliced}. This could be applied after the learning phase of {\KNIFE} and is left for future work.

Another promising direction for future research is the possible combination of different estimation methods and the resulting bias-variance tradeoff.

\section*{Acknowledgments}
This work was granted access to the HPC resources of IDRIS under the allocation 2021-101838 made by GENCI. 
G.\ K.\ gratefully acknowledges support from the Austrian Science Fund
(FWF): Y 1199.

\clearpage\clearpage \bibliographystyle{icml2022}
\bibliography{IEEEabrv.bib,literature.bib}

\newpage
\clearpage
\appendix
\onecolumn
~
\begin{center}
  \vspace{-.5\baselineskip}
  {\LARGE \textsc{Appendix}}  
\end{center}

\section{Experimental Details of Experiments with Synthetic Data}
\label{sec:exper-deta-exper}

Implementation of {\KNIFE} in PyTorch~\citep{pytorch} is rather straightforward. The constraint on the weights $\vt u$ can be satisfied by applying a \texttt{softmax} transformation. The covariance matrices were parameterized by the lower-triangular factor in the Cholesky decomposition of the precision matrices, guaranteeing the definiteness constraint to be satisfied.

\subsection{Differential Entropy Estimation of Gaussian Data}
\label{sec:appendix:gaussian-data}

In \cref{sec:gauss}, the estimation of the entropy $\Ent(X) = \frac{d}{2} \log 2\pi e$ for $X \sim \mathcal N(\vt 0, I_d)$ was performed with the hyperparameters given in \cref{tab:ib_details:gauss1}. The mean error and its empirical standard deviation are reported in \cref{tab:gaussian_results} over $20$ runs, where an independently drawn evaluation set with the same size as the training set is used. At $d=10$ we have the entropy \pgfmathparse{10/2*ln(2*pi*exp(1))}$h = \frac{d}{2} \log 2\pi e = \pgfmathprintnumber{\pgfmathresult}$, while for the higher dimension, $d=64$ we find \pgfmathparse{64/2*ln(2*pi*exp(1))}$h = \pgfmathprintnumber{\pgfmathresult}$.

In the experiment depicted in \cref{fig:gauss}, entropy is decreased after every epoch by letting $X_i \sim \mathcal N(\vt 0, a^i I_d)$, where $i=0,\dots,4$ is the epoch index. That is, $X_i = \sqrt{a^i} G^d$, where $G$ is a standard normal random variable, resulting in an decrease of the DE by $\Delta = -\frac{d}{2}\log a \approx \pgfmathparse{64/2*ln(2)}\pgfmathprintnumber{\pgfmathresult}$ for $a = \frac{1}{2}$ with every epoch. We start at \pgfmathparse{64/2*ln(2*pi*exp(1))}$\Ent(X_0) = \frac{d}{2} \log{2\pi e} \approx \pgfmathprintnumber{\pgfmathresult}$ and successively decrease until \pgfmathparse{64/2*ln(2*pi*exp(1))-2*64*ln(2)}$\Ent(X_4) = \Ent(X_0) + 4\Delta \approx \pgfmathprintnumber{\pgfmathresult}$. Additional parameters can be found in \cref{tab:ib_details:gauss2}.

\begin{table}[b]
  \centering
  \begin{minipage}{.47\textwidth}
    \centering
    \caption{Experimental details of first experiment in \cref{sec:gauss}.}\vspace{1mm}
    \label{tab:ib_details:gauss1}
    \begin{tabular}{cc}
    \toprule
      Parameter       & Value  \\
      \midrule
      Source Distribution $X$ & $X \sim \mathcal N(\vt 0,I_d)$ \\
      Dimension $d$   & $10$ and $64$  \\
      Optimizer       & Adam \\
      Learning Rate   & 0.01\\
      Batch Size $N$  & 128 \\
      Kernel Size $M$ & 128 \\
      Iterations per epoch & 200 \\
      Epochs          & 1 \\
      Runs            & 20 \\
      \bottomrule
    \end{tabular}
  \end{minipage}\hfill
  \begin{minipage}{.47\textwidth}   
    \centering
    \caption{Experimental details of the experiment depicted in \cref{fig:gauss}.}\vspace{1mm}
    \label{tab:ib_details:gauss2}
    \begin{tabular}{cc}
    \toprule
      Parameter       & Value  \\
      \midrule
      Source Distribution $X$ & $X \sim \mathcal N(\vt 0,a^i I_d)$ \\
                      & for $i=0,\dots,4$\\
      Dimension $d$   & $64$  \\
      Factor $a$      & $\frac{1}{2}$  \\
      Optimizer       & Adam \\
      Learning Rate   & 0.01\\
      Batch Size $N$  & 128 \\
      Kernel Size $M$ & 128 \\
      Iterations per epoch & 1000 \\
      Epochs          & 5 \\
      Runs            & 1 \\
      \bottomrule
    \end{tabular}
  \end{minipage}
\end{table}

\paragraph{Computational Resources.}
Training was performed on an NVidia V100 GPU. Taken together, training for the first experiments of entropy estimation in dimensions $d=10,64$, as well as the experiment depicted in \cref{fig:gauss} used GPU time of less than $5$ minutes.

\subsection{Differential Entropy Estimation of Triangle Mixtures}
\label{sec:triangle-mixtures}

In \cref{sec:ent_triangle}, we perform an estimation of the entropy of $c$-component triangle mixture distributions.
The PDF of such a $c$-component triangle-mixture, is given by
\begin{equation}
  \label{eq:1}
  p(x) = \sum_{i=1}^c w_i \Lambda_{s_i}\bigg(x-i-\frac{1}{2}\bigg) ,
\end{equation}
where $\Lambda_s(x) := \frac{1}{s} \max\{0,2-4s|x|\}$ is a centered triangle PDF with width $s>0$. The scales $\vt s = (s_1,\dots,s_c)$ and weights $\vt w = (w_1,\dots,w_c)$ satisfy $0 < s_i, w_i < 1$ and $\sum_{i=1}^c w_i = 1$.
Before the experiment, we choose $\vt w$ uniformly at random from the $c$-probability simplex and the scales are chosen uniformly at random in $[0.1, 1.0]$. An example for $c=10$ is the true PDF depicted in \cref{fig:tirangle_mixture_expe} (left). For $d > 1$, we perform the estimation on $d$ i.i.d.\ copies. Note that the triangle mixture with $c$ components in $d$-dimensional space has $c^d$ modes, i.e., the support can be partitioned into $c^d$ disjoint components.

The parameters of the experiment yielding \cref{fig:tirangle_mixture_expe} (left) are given in \cref{tab:ib_details:trimix}, while the details of the experiment depicted in \cref{fig:tirangle_mixture_expe} (right) can be found in \cref{tab:ib_details:trimix2}. In the latter experiment, over ten runs, entropy was estimated to an accuracy of $1.6563 \pm 0.8528$ by {\KNIFE}, accurate to $2.4445 \pm 0.5439$ using \cref{eq:schraudolph} and with an accuracy of $7.1070 \pm 2.7984$ by {\doe}. This is the mean absolute error and its empirical standard deviation over all $10$ runs, where the evaluation set was drawn independently from the training set and has the same size as the training set.

\begin{table}
  \centering
  \begin{minipage}{.47\textwidth}
    \centering
     \caption{Experimental details of the experiment resulting in the PDF in \cref{fig:tirangle_mixture_expe} (left).}\vspace{1mm}
    \label{tab:ib_details:trimix}
    \begin{tabular}{cc}
    \toprule
      Parameter       & Value  \\
      \midrule
      Source Distribution $X$ & $c$-component \\
                      & triangle mixtures \\
      Components $c$  & $10$ \\
      Dimension $d$   & $1$  \\
      Optimizer       & Adam \\
      Learning Rate   & 0.1\\
      Batch Size $N$  & 128 \\
      Kernel Size $M$ & 128 \\
      Iterations per epoch & 100 \\
      Epochs          & 10 \\
      Runs            & 1 \\
    \bottomrule
    \end{tabular}
   
  \end{minipage}\hfill
  \begin{minipage}{.47\textwidth}    \centering
  \caption{Experimental details of the experiment resulting in the training depicted in \cref{fig:tirangle_mixture_expe} (right).}\vspace{1mm}
    \label{tab:ib_details:trimix2}
    \begin{tabular}{cc}
    \toprule
      Parameter       & Value  \\
      \midrule
      Source Distribution $X$ & $c$-component \\
                      & triangle mixtures \\
      Components $c$  & $2$ \\
      Dimension $d$   & $8$  \\
      Optimizer       & Adam \\
      Learning Rate   & 0.001\\
      Batch Size $N$  & 128 \\
      Kernel Size $M$ & 128 \\
      Iterations per epoch & 1000 \\
      Epochs          & 20 \\
      Runs            & 10 \\
      \bottomrule
    \end{tabular}
  \end{minipage}
\end{table}

\paragraph{Computational Resources.}
Training was performed on an NVidia V100 GPU. Training in $d=1$ dimension, that resulted in \cref{fig:tirangle_mixture_expe}~(left) can be performed in seconds, while all training required for producing \cref{fig:tirangle_mixture_expe}~(right) used approximately $1.5$ hours of GPU time.

\subsection{Mutual Information Estimation}
\label{sec:app:MI_estimation}
In \cref{sec:synthetic_mi_estimation}, we estimate $\MI(X^d;Y^d)$ and $\MI(X^d;(Y^3)^d)$ where $(X,Y)$ are multivariate correlated Gaussian distributions with correlation coefficient $\rho_i$ in the $i$-th epoch.
Subsequently, we estimate $\MI(X^d;Y^d)$ where $X,E \sim \mathcal U[-\sqrt{3}, \sqrt{3}]$ are independent and $Y$ is given by $Y = \rho_i X + \sqrt{1 - \rho_i^2} E$.
In both cases, $\rho_i$ is chosen such that $\MI(X^d;Y^d) = 2i$ in the $i$-th epoch.

\begin{figure*}
  \centering
  \begin{minipage}{0.5\textwidth}
    \centering
    \includegraphics[width=\textwidth]{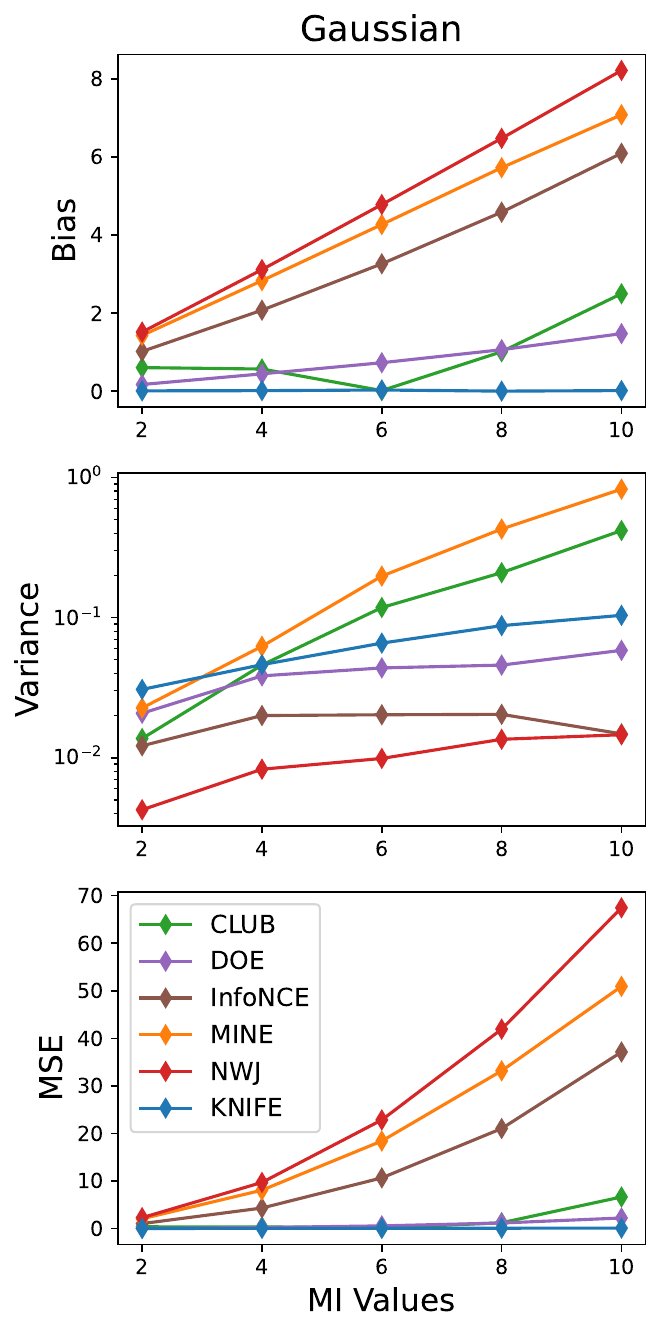}
  \end{minipage}%
  \begin{minipage}{0.5\textwidth}
    \centering
    \includegraphics[width=\textwidth]{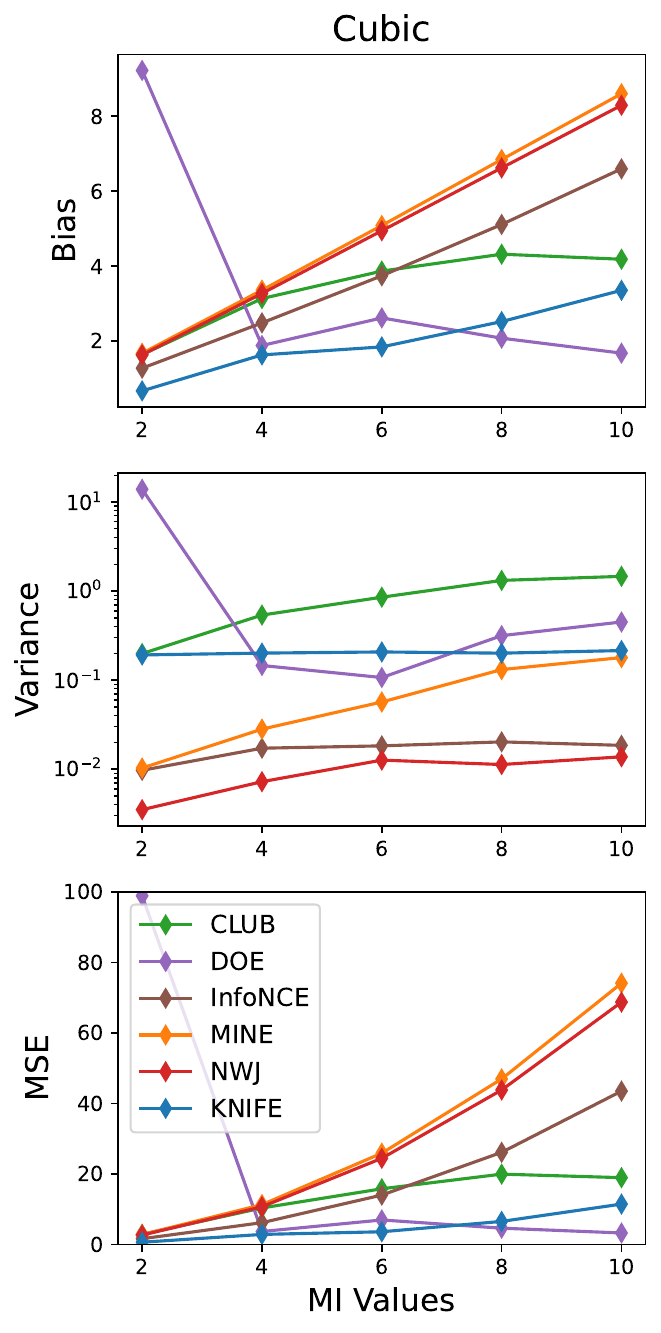}
  \end{minipage}
  \caption{\textbf{Left}: Estimation of $\MI(X;Y)$; \textbf{Right}: Estimation of $\MI(X; Y^3)$ (cubic transformation).}\label{fig:appendix:gauss}
\end{figure*}

All neural networks are randomly initialized. The bias, variance, and MSE during training as a function of the MI, can be observed in \cref{fig:appendix:gauss}.

\begin{figure}
  \centering
  \includegraphics[width=.5\textwidth]{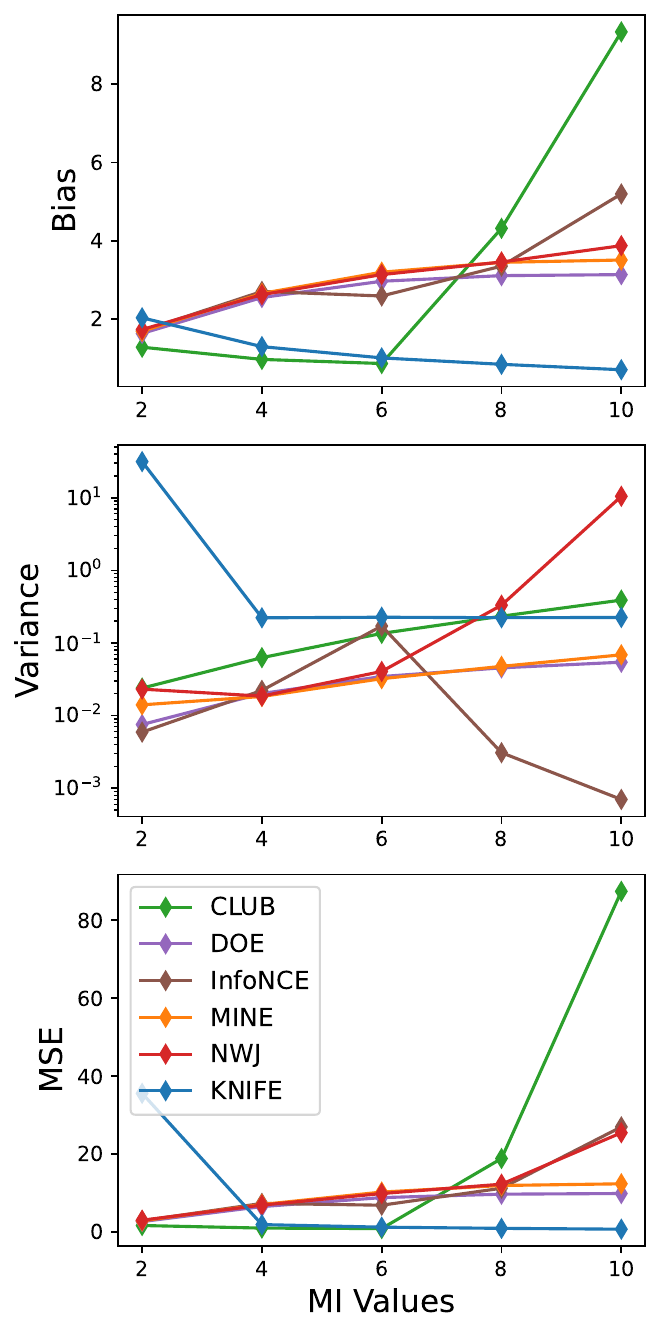}
  \caption{Bias, variance, and MSE for MI estimation on uniformly distributed data.}\label{fig:appendix:uniform_bias_var_mse}
\end{figure}

The estimation is performed in $10$ runs, randomly choosing the training meta-parameters as proposed by \citep{pmlr-v108-mcallester20a}. 
In \cref{fig:mi_gauss}, in light blue, we show the best run for each method by distance from the true MI at the end of training. An exponentially weighted average is shown in dark blue. The bias, variance, and MSE during training, as a function of the MI, can be observed in \cref{fig:appendix:uniform_bias_var_mse}.
Details about the source distribution as well as details of the experiments can be found in \cref{tab:appendix:mi_uniform}. During experimentation it turned out to be beneficial to train the parameters $\bs\Theta$ and $\bs\theta$ in \cref{eq:MI} separately and substantially increase the learning rate for the training of $\bs\theta$. Thus, we increase the learning rate for the training of $\bs\theta$ by a factor of $10^3$.

\begin{figure}[h]
    \centering
    \includegraphics[width=.45\textwidth]{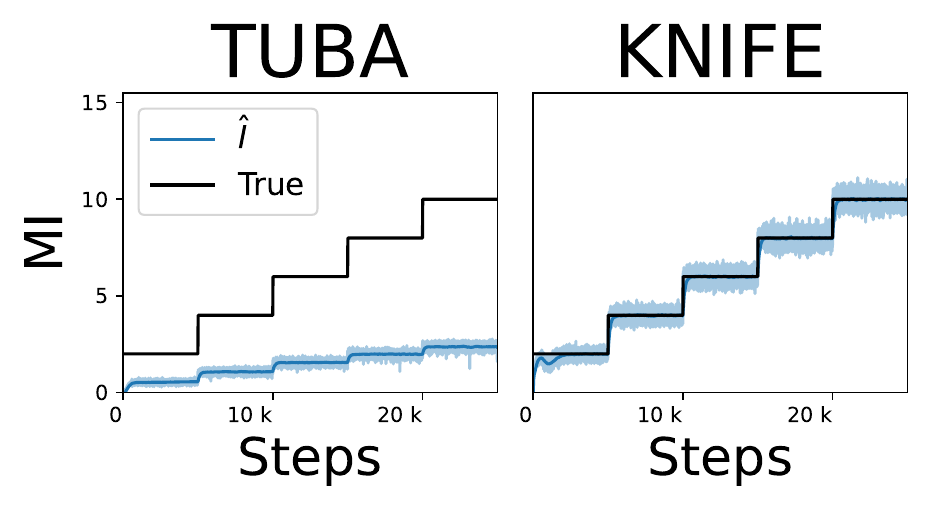}
    \includegraphics[width=.45\textwidth]{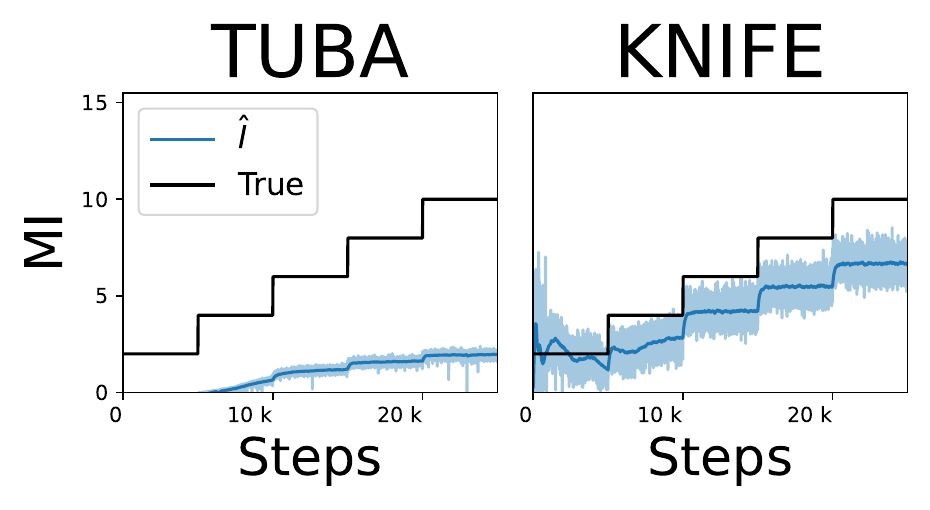}
    \caption{Estimation of $I(X^d;Y^d)$ \textbf{(left)} and $I(X^d;(Y^3)^d)$ \textbf{(right)} as in \cref{fig:mi_gauss}.}
    \label{fig:tuba}
\end{figure}

\Cref{fig:tuba} depicts the same experiments, repeated for {\TUBA}~\cite{pmlr-v97-poole19a} which is a tractable version of the Barber-Agakov bound. It is worth noting that {\NWJ} is a special case of {\TUBA}.

\paragraph{Model Architecture for $\bs\Theta$.} We utilize the feed-forward architecture, also used in \citep{pmlr-v108-mcallester20a}. It is a simple architecture with two linear layers, one hidden layer using $\tanh$ activation, immediately followed by an output layer. The number of neurons in the hidden layer is a meta-parameter selected randomly from $\{64, 128, 256\}$ for each training run.
Three models with this architecture are used for the three parameters $(\vt A, \vt b, \vt u)$, as described by \cref{eq:knife}, where only the output dimension is changed to fit the parameter dimension.

\begin{table*}
  \centering
  \begin{minipage}{1\textwidth}
    \centering
        \caption{Experimental details of the training depicted in \cref{fig:mi_gauss} (bottom).}\vspace{1mm}
    \begin{tabular}{cc}
    \toprule
      Parameter       & Value  \\
      \midrule
      Dimension $d$   & $20$  \\
      Optimizer       & Adam \\
      Learning Rates  & 0.01, 0.003, 0.001, 0.0003 \\
      Batch Size $N$  & 128 \\
      Kernel Size $M$ & 128 \\
      Iterations per epoch & 25\,000 \\
      Epochs          & 1 \\
      Runs            & 10 \\
      \bottomrule
    \end{tabular}
    \label{tab:appendix:mi_uniform}
  \end{minipage}
\end{table*}

\paragraph{Computational Resources.}
Training was performed, using about $6$ hours of GPU time on an NVidia V100 GPU to carry out the experiment depicted in \cref{fig:mi_gauss}~(bottom).

\section{Additional Experiments on Synthetic Data}
\label{sec:addit-exper-synth}

\subsection{Entropy Estimation of Triangle Mixtures}
\label{sec:entr-estim-triangle}

\begin{figure}[h]
  \centering
  \includegraphics[width=.6\linewidth]{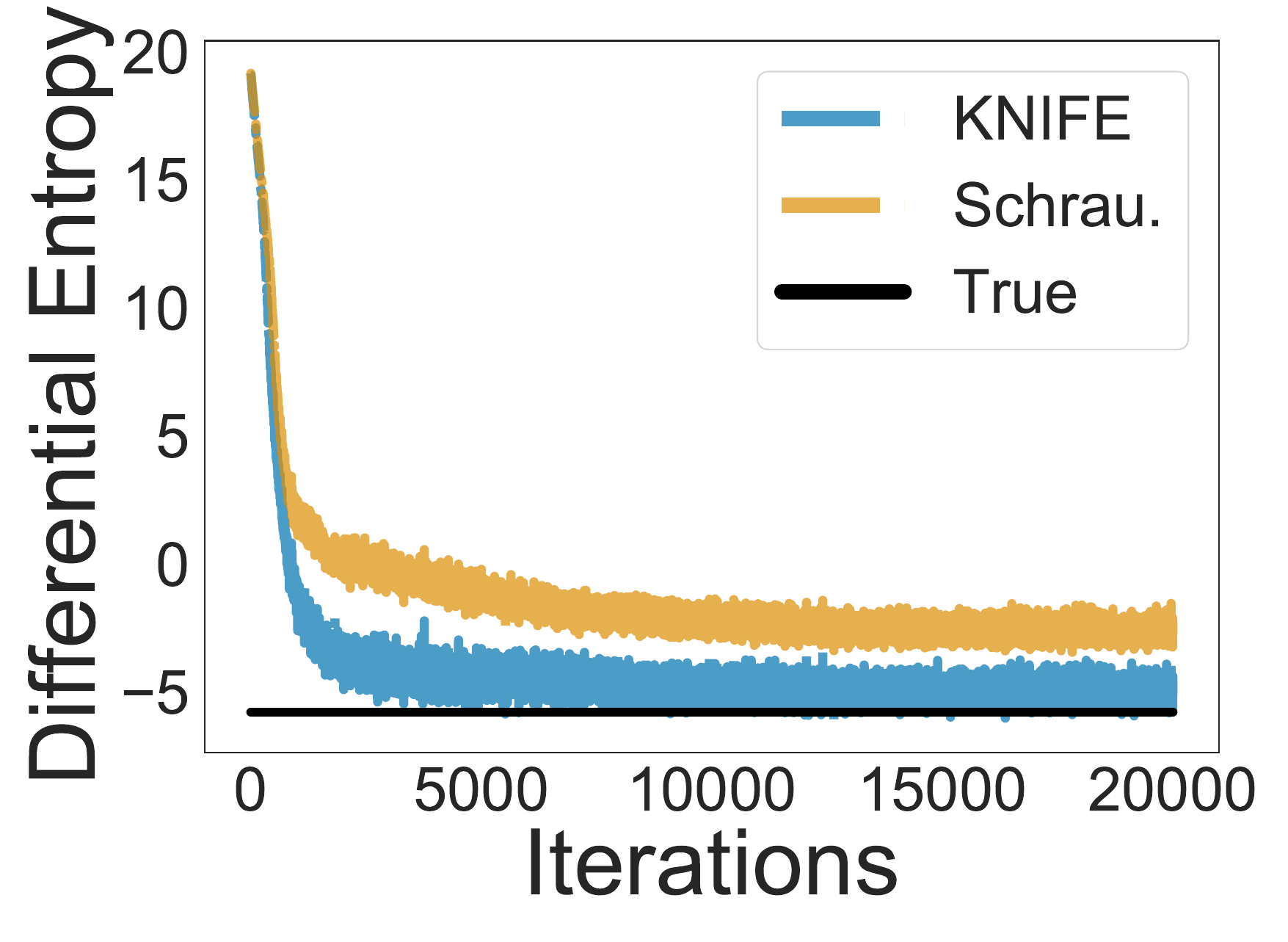}
  \caption{Entropy estimation of a $2$-component triangle mixture in $d=20$ dimensions.}
  \label{fig:app:triangle_d-20}
\end{figure}

In addition to the experiments performed in \cref{sec:ent_triangle}, we repeated the estimation of DE of a $c$-component triangle mixture in $d=20$ dimensions with $c=2$ components. In this more challenging setting, we found that over ten runs, DE was estimated to an accuracy of $10.6270 \pm 8.2179$ by {\KNIFE} and accurate to $13.1660 \pm 8.2651$ using \cref{eq:schraudolph}. The mean error and standard deviation is computed using and independently drawn evaluation set that has the same size as the training set.
The best training run is depicted in \cref{fig:app:triangle_d-20} and the details can be found in \cref{tab:app:trimix_d-20}.

\begin{table}[h]
  \centering
    \caption{Experimental details of the experiment resulting in the training depicted in \cref{fig:app:triangle_d-20}.}\vspace{1mm}
  \begin{tabular}{cc}
  \toprule
    Parameter       & Value  \\
    \midrule
    Source Distribution $X$ & $c$-component triangle mixtures \\
    Components $c$  & $2$ \\
    Dimension $d$   & $20$  \\
    Optimizer       & Adam~\citep{Kingma2014Adam} \\
    Learning Rate   & 0.001\\
    Batch Size $N$  & 128 \\
    Kernel Size $M$ & 128 \\
    Iterations per epoch & 1000 \\
    Epochs          & 20 \\
    Runs            & 10 \\
    \bottomrule
  \end{tabular}
  \label{tab:app:trimix_d-20}
\end{table}

\section{Experimental Details of Experiments on Natural Data}
\label{sec:expdetailssupp}

\subsection{On the parameter update}

In \cref{sec:natural_data}, we rely on two different types of models: pretrained (e.g.,  fine tuning with VIBERT) and randomly initialized (e.g.,  in fair classification and domain adaptation). When working with randomly initialized networks the parameters are updated. However, it is worth noting that in the literature the pretrained model parameters (\textit{i.e.}\ $\psi$) are not always updated (see \citep{ravfogel2020null}). 
In our experiments: (i) We always update the parameters (even for pretrained models), and (ii) we did not change the way the parameters were updated in concurrent works (to ensure fair comparison). Specifically,
\begin{itemize}
    \item for language model finetuning (\cref{ssec:ib_finetuning_lm}), we followed \citep{mahabadi2021variational} and did a joint update;
    \item for the fair classification task (\cref{ssec:fair_classif_sup}), we followed common practice and used the algorithm described in \cref{alg:fitted_disent2} which rely on an alternated update;
    \item for the domain adaptation task (\cref{ssec:unsupervized}), we followed common practice and used a joint method.
\end{itemize}

\begin{algorithm}
  \caption{{Disentanglement using a MI-based regularizer
    }}
  \begin{algorithmic}[1]
    \State \textsc{Input}  Labeled training set $\mathcal{D}=(x_{n},s_{n},y_{n})_{n = 1,\dots, N}$; independent set of samples $\Dm$; parameters of {\KNIFE} $\theta$; parameters of network $\psi$.
    \State \textsc{Initialize} parameters $\theta$, $\psi$
    \State \textsc{Optimization}
    \While{$(\theta,\psi)$ not converged} 
      \For{\texttt{ $k \in \{1,\cdots,K\}$}} \Comment{Learning Step for {\KNIFE}}
        \State Sample a batch $\mathcal{B}$ from $\Dm$
        \State Update $\theta$ using \cref{eq:MI}.
      \EndFor
      \State Sample a batch $\mathcal{B^\prime}$ from $\mathcal{D}$
      \State Update $\theta$ with $\mathcal{B^\prime}$ \cref{eq:all_loss}.
    \EndWhile
    \State \textsc{Output}  Encoder and classifier weights ${\psi}$
  \end{algorithmic}
  \label{alg:fitted_disent2}
\end{algorithm}

\subsection{Information Bottleneck for Language Model Finetuning}\label{ssec:ib_finetuning_lm}
For this experiment we follow the experimental setting introduced in  \citep{mahabadi2021variational} and work with the GLUE data\footnote{see \url{https://gluebenchmark.com/faq}}.
\paragraph{Model Architecture.} We report in \cref{tab:architecture_vibert}, the multilayer perceptron (MLP) used to compute the compressed sentence representations produced by BERT. Variance and Mean MLP networks are composed of fully connected layers.
\begin{table*}
  \begin{minipage}{0.45\textwidth}
      \caption{Architecture of the model used in the IB finetuning experiment. We use ReLU       as an activation function.}\vspace{1mm}
    \label{tab:architecture_vibert}
    \begin{tabular}{ccc}
    \toprule
      Layer type  &  Input shape & Output shape \\
      \midrule
      Fully connected  & 768  & $\frac{2304+K}{4}$ \\
      Fully connected  & $\frac{2304+K}{4}$ & $\frac{768+K}{2}$ \\ 
      \bottomrule
    \end{tabular}
  \end{minipage}\hfill
  \begin{minipage}{0.45\textwidth}
      \caption{Experimental details on Information Bottleneck.}\vspace{1mm}
    \label{tab:ib_classif_details}
    \centering
    \begin{tabular}{cc}
    \toprule
      Parameter & Value  \\
      \midrule
      Learning Rate     & See \cref{ssec:ib_finetuning_lm}  \\
      Optimizer     & AdamW       \\
      Warmup Steps     & 0.0 \\
      Dropout     & 0.0 \\
      Batch Size     & 32 \\
      \bottomrule
    \end{tabular}
  \end{minipage} 
\end{table*}

\paragraph{Model Training.} For model training, all models are trained for 6 epochs and we use early stopping (best model is selected on validation set error). For IB, $\lambda$ is selected in $\{10^{-4},10^{-5},10^{-6}\}$ and $K$ is selected in $\{144, 192, 288, 384\}$. 
We follow \citep{DBLP:journals/corr/AlemiFD016} where the posterior is averaged over 5 samples and a linear annealing schedule is used for $\lambda$. Additional hyper-parameters are reported in \cref{tab:ib_classif_details}.
\paragraph{Dataset Statistics.} \cref{tab:glue_Stats} reports the statistics of the dataset used in our finetuning experiment.
\paragraph{Computational Resources.} For all these experiments we rely on NVidia-P100 with 16GB of RAM. To complete the full grid-search on 10 seeds and on the three datasets, approximately 1.5k hours are required.

\begin{table}
\centering
    \caption{Datasets from the GLUE as used in our experiments.}\vspace{1mm}
    \label{tab:glue_Stats}
    \begin{tabular}{ccccc}
    \toprule
         & \#Labels& Train &Val. &Test \\ 
    \midrule
      RTE&  2&  2.5k & 0.08k&  3k  \\
      STS-B&  1 (regression) & 5.8k&  1.5k & 1.4k  \\
      MRPC&  2 & 3.7k&  0.4k&  1.7k  \\
      \bottomrule
    \end{tabular}

\end{table}

\subsection{Fair Textual Classification}\label{ssec:fair_classif_sup}
In this section, we gather the experimental details for the textual fair classification task.

\subsubsection{Details of the \KNIFE-based Estimator}

In this experiment, we estimate the MI between a continuous random variable, namely $Z = \Phi_\psi(X)$, and a discrete variable, denoted by $S \in \mathcal{S}=\{1,2,\dots,|\mathcal{S}| \}$.
We follow the strategy outlined in \cref{sec:cond_mi} for estimating the conditional DE $\Ent(Z|S)$. 
However, we will reuse the estimate of the conditional PDF $\hat p(z|s;\bs\Theta)$ to compute an estimate of the DE as
\begin{align}\label{eq:mi_knife_doe}
\Ent(Z)      &\approx -\frac{1}{N} \sum_{n=1}^N \log \left( \sum_{s \in \mathcal{S}} \hat p_{\KNIFE}(z_n|s; \bs\Theta) \hat{p}(s) \right),
\end{align}
where $\hat{p}(s) = \frac{1}{N} |\{n : s_n = s\}|$ is used to indicate the empirical distribution of $S$ in the training set $\Es$.\footnote{As we work with balanced batches, we will have  $\hat{p}(s)=\frac{1}{|\mathcal{S}|}$.} 
In our experiments, with $|\mathcal{S}| = 2$, we found that estimating the DE $\Ent(Z)$ based on the {\KNIFE} estimator learnt for $\Ent(Z|S)$ increases the stability of training. We adopted the same strategy for {\doe}.

\subsubsection{Experimental Details}
\paragraph{Model Architecture.} For the encoder, we use a bidirectional GRU with two layers with hidden and input dimension set to 128. We use LeakyReLU as the activation function. The classification head is composed of fully connected layers of input dimension 256. We use a learning rate of $0.0001$ for AdamW. The dropout rate is set to 0.2. The number of warmup steps 
is set to 1000. 
\paragraph{Computational Resources.} For all these experiments, we rely on NVIDIA-P100 with 16GB of RAM. Each model is trained for 30k steps. The model with the lowest MI is selected. The training of a single network takes around 3 hours. 

\subsection{Unsupervised Domain Adaptation}\label{ssec:unsupervized}

We follow the experimental setup given in \citep{Cheng2020CLUB} as closely as possible, i.e., we pick hyperparameters given in the paper, or if not provided, those set in the code:\footnote{\url{https://github.com/Linear95/CLUB/tree/master/MI_DA}.}

\paragraph{Model Training.}
    We use Adam optimizer for all modules with a learning rate of 0.001. Batch size is set to 128. We set the weighting parameter $\lambda=0.1$. The original code of \citep{Cheng2020CLUB} uses 15\,000 training iterations, but we found most methods had not properly converged at this stage, and hence use 25\,000 iterations instead. Similar to other experiments, we set the kernel size $M=128$.

\paragraph{Model Architecture.} \Cref{tab:da_architectures} summarizes the architectures used for the different modules. For the MI network of each method, the best configuration, based on the validation set of the first task MNIST $\rightarrow$ MNIST-M, is chosen among 4 configurations: with or without LayerNorm and with ReLU or $\tanh$ activation.

\paragraph{Computational Resources.} For these experiments, we used a cluster of NVIDIA-V100 with 16GB of RAM. Each training (i.e., 25k iterations) on a single task requires on average 2 hours. Given that we have 6 tasks, and repeat the training for 3 different seeds, on average 36 hours computation time is required for each method.

\begin{table*}[h]
  \centering
  \caption{Architectures used for the Unsupervised Domain Adaptation experiments. For the MI network of each method, we chose the best performing configuration between with or without LayerNorm layer and best activation between ReLU and $\tanh$, using the validation set of MNIST-M.}\vspace{1mm}
  \begin{tabular}{cccc}
    \multicolumn{4}{c}{\textbf{Encoder} (both $\Phi$ and $\Phi^d$)} \\
    \toprule
    Layer type &  Input shape & Output shape & Details \\
    \midrule
    Convolution sequence & (3, H, W) & (64, H, W) & Cf. below \\
    Noisy downsampling & (64, H, W) & (64, H // 2, W // 2) & Cf. below \\
    Convolution sequence & (64, H // 2, W // 2) & (64, H // 2, W // 2) & Cf. below \\
    Noisy downsampling & (64, H // 2, W // 2) & (64, H // 4, W // 4) & Cf. below \\
    Convolution sequence & (64, H // 4, W // 4) & (64, H // 4, W // 4) & Cf. below \\
    Global Average Pool & (64, H // 4, W // 4) & (64,) & - \\
    \bottomrule
  \end{tabular} \\
  \vspace{1em}
  \begin{minipage}{0.48\linewidth}
    \resizebox{\textwidth}{!}{
      \begin{tabular}{ccc}
        \multicolumn{3}{c}{\textbf{Main classifier} $C$} \\
        \toprule
        Layer type &  Input shape & Output shape \\
        \midrule
        Fully connected & (64,) & (10,) \\
        \bottomrule
      \end{tabular}
    }
  \end{minipage}
  \begin{minipage}{0.48\linewidth}
    \resizebox{\textwidth}{!}{
      \begin{tabular}{ccc}
        \multicolumn{3}{c}{\textbf{Domain classifier} $C^d$} \\
        \toprule
        Layer type &  Input shape & Output shape \\
        \midrule
        Fully connected & (64,) & (2,) \\
        \bottomrule
      \end{tabular}
    }
  \end{minipage}

  \vspace{1em}
  \begin{tabular}{cccc}
    \multicolumn{4}{c}{\textbf{Convolution sequence}} \\
    \toprule
    Layer type &  Input shape & Output shape & Parameters \\
    \midrule
    2D convolution & (3, H, W) & (64, H, W) & 3x3, 64 channels, Stride=1, Padding=1 \\
    2D BatchNorm   & (3, H, W) & (64, H, W) & - \\
    Activation & (3, H, W) & (64, H, W) & LeakyReLU 0.1 \\
    2D convolution & (64, H, W) & (64, H, W) & 3x3, 64 channels, Stride=1, Padding=1 \\
    2D BatchNorm   & (64, H, W) & (64, H, W) & - \\
    Activation & (64, H, W) & (64, H, W) & LeakyReLU 0.1 \\
    2D convolution & (64, H, W) & (64, H, W) & 3x3, 64 channels, Stride=1, Padding=1 \\
    2D BatchNorm   & (64, H, W) & (64, H, W) & - \\
    Activation & (64, H, W) & (64, H, W) & LeakyReLU 0.1 \\
    \bottomrule
  \end{tabular} \\
  \vspace{1em}
  \begin{tabular}{cccc}
    \multicolumn{4}{c}{\textbf{Noisy downsampling}} \\
    \toprule
    Layer type &  Input shape & Output shape & Parameters \\
    \midrule
    MaxPool & (64, H, W) & (64, H // 2, H // 2) & 2x2, Stride=2 \\
    Dropout & (64, H // 2, W // 2) & (64, H // 2, H // 2) & p=0.5 \\
    Noise & (64, H // 2, W // 2) & (64, H // 2, H // 2) & Gaussian with $\sigma=1$ \\
    \bottomrule
  \end{tabular}\\
  \vspace{1em}
  \begin{tabular}{cccc}
    \multicolumn{4}{c}{\textbf{MI network}} \\
    \toprule
    Layer type &  Input shape & Output shape & Details \\
    \midrule
    LayerNorm & ($C_{in}$,) & ($C_{in}$,) & Optional \\
    Fully connected & ($C_{in}$,) & (64,) & Activation = [ReLU, $\tanh$] \\
    LayerNorm & ($C_{in}$,) & (64,) & Optional \\
    Fully connected & (64,) & ($C_{out}$,) & Optional \\
    \bottomrule
  \end{tabular}
  \label{tab:da_architectures}
\end{table*}

\section{Bounding the Error}
\label{sec:error-bound}

In the following, fix $L > 0$ and let $\PPP_L$ be the set of $L$-Lipschitz PDFs supported\footnote{Any known compact support suffices. An affine transformation then yields $\XXX = [0,1]^d$, while possibly resulting in a different Lipschitz constant.} on $\XXX \defas [0,1]^d$, \ie, $\int_{\XXX} p(x) \,\dd x = 1$, and 
\begin{align}
  \forall  x,  y \in \RR^d : |p(x) - p(y)| \le L \norm{x - y}
\end{align}
for $p \in \PPP_L$, where\footnote{The $\ell_1$ norm is chosen to facilitate subsequent computations. By the equivalence of norms on $\RR^d$, any norm suffices.} $\norm{x} \defas \sum_k{|x_k|}$.

Assume $p \in \PPP_L$ and let $\kernel$ be a PDF supported on $\XXX$. In order to show that estimation of $\Ent(X)$ is achievable, we use a standard Parzen-Rosenblatt estimator $\htp{x; w} \defas \frac{1}{M w^d}\sum_{m=1}^{M} \kernel\big(\frac{x - X'_m}{w}\big)$, as in~\cref{eq:kernel0}.
The entropy estimate is then defined by the empirical average
\begin{align}
  \label{eq:12}
  \widehat \Ent(\Dn; w) \defas - \frac1N \sum_{n=1}^N \log \htp{X_n; w} .
\end{align}
Further, define the following quantities, which are assumed to be finite:
\begin{align}
  p_{\mathrm{max}} &\defas \max\{p(x): x \in \XXX\}, \\
    C_1 &\defas \int p(x) \log^2 p(x) \dd x, \\
  C_2 &\defas L  \int \norm{u} \kernel(u) \dd u, \\
  K_{\mathrm{max}} &\defas \max\{\kernel(x): x \in \XXX\} .
\end{align}
Note that it is easily seen that $p_{\mathrm{max}} \le \frac{L}{2}$ and $C_1 \le \max\big\{p_{\mathrm{max}} \log^2 p_{\mathrm{max}}, 4e^{-2}\big\}$ by our assumptions. The requirement $C_2, K_{\mathrm{max}} < \infty$ represents a mild condition on the kernel function $\kappa$.

We can now show the following.
\begin{theorem}
  \label{thm:confidence_bound}
  With probability greater than $1-\delta$ we have
  \begin{align}
    |\Ent(X) - \widehat \Ent(\Dn; w)|  \le  -\log  \left( 1 - \frac{3 N K_{\mathrm{max}}}{w^d \delta} \sqrt{\frac{\log\frac{6N}{\delta}}{2M}} - \frac{3 N C_2 w}{\delta} \right) + \sqrt{\frac{3 C_1}{N \delta}}  ,
  \end{align}
  if the expression in the logarithm is positive.
\end{theorem}

In particular, the estimation error approaches zero as $N \to \infty$ if $w = w(N)\to 0$, $M = M(N)\to \infty$ are chosen such that
\begin{align}
  Nw &\to 0, \\
  \frac{N^2 \log N}{w^{2d} M} &\to 0 .
\end{align}

We prove \cref{thm:confidence_bound} in several Lemmas.
\begin{lemma}
  \label{lem:p_phat}
  Fix $\delta > 0$ and $x_0 \in \XXX$. Then, with probability greater than $1-\delta$, 
  \begin{align}
    \label{eq:14}
    |p(x_0) - \htp{x_0}| \le \frac{K_{\mathrm{max}}}{w^d} \sqrt{\frac{\log\frac{2}{\delta}}{2M}} + C_2 w .
  \end{align}
\end{lemma}
\begin{proof}
  First, we can show that
  \begin{align}
    |\Exp{\htp{x_0}}&  - p(x_0)| \\ & = \left| \frac{1}{M w^d}\sum_{m=1}^{M} \int \kernel\left(\frac{x_0 - x}{w}\right) p(x) \dd x  - p(x_0) \right| \\
                                       &= \left| \frac{1}{w^d} \int \kernel\left(\frac{x_0 - x}{w}\right) p(x) \dd x  - p(x_0) \right| \\
                                       &= \left| \int \kernel\left(u \right) p(x_0 - w u) \dd u  - p(x_0) \right| \\
                                       &= \left| \int \kernel\left(u \right) [p(x_0 - w u)   - p(x_0)] \dd u\right| \\
                                       &\le \int \kernel\left(u \right) | p(x_0 - w u)   - p(x_0) | \dd u \\
                                       &\le \int \kernel\left(u \right) L w \Vert u \Vert  \dd u \\
                                       &= w C_2 . \label{eq:kernel_expectations}
  \end{align}
    Next, note that
  \begin{align}
    \label{eq:mcdiarmid_approx}
    |\Exp{\htp{x_0}} - \htp{x_0}| \le \frac{K_{\mathrm{max}}}{w^d} \sqrt{\frac{\log\frac{2}{\delta}}{2M}}
  \end{align}
  holds with probability greater than $1-\delta$ as the requirements of McDiarmid's inequality \citep[Sec.~3]{Paninski2003Estimation} are satisfied with $c_j = \frac{K_{\mathrm{max}}}{M w^d}$ and thus $\Prob{|\Exp{\htp{x_0}} - \htp{x_0}| \ge \eps} \le \delta$
  with
  \begin{align}
        \eps &=            \frac{K_{\mathrm{max}}}{w^d}  \sqrt{\frac{\log\frac{2}{\delta}}{2M}} .
  \end{align}
  Combining \cref{eq:kernel_expectations} and \cref{eq:mcdiarmid_approx} gives \cref{eq:14}.
\end{proof}

\begin{lemma}
  \label{lem:prob_p}
  For any continuous random variable $X$ supported on $\XXX$ and $a \ge 0$, we have
  \begin{align}
    \label{eq:6}
    \Prob{p(X) \le a} \le a .
  \end{align}
\end{lemma}
\begin{proof}
  We apply Markov's inequality to the random variable $Y = \frac{1}{p(X)}$ and observe that
  \begin{align}
    \label{eq:7}
    \Prob{p(X) \le a} = \Prob{Y \ge a^{-1}} \le \vol(\XXX) a  = a.
  \end{align}
\end{proof}

\begin{lemma}
  \label{lem:log_ineq}
  If $x > 0$, $y \ge a > 0$, $0 < a < 1$, and $|x-y| \le \delta < a$, then
  \begin{align}
    \label{eq:8}
    |\log x - \log y| \le \log\frac{a}{a-\delta} = -\log\left(1 - \frac{\delta}{a}\right) .
  \end{align}
\end{lemma}
\begin{proof}
  \textbf{Case $x \ge y$.} We can write $y = a + b$ and $x = y + c = a + b + c$ for $b \ge 0$ and $0 \le c \le \delta < a$.
  \begin{align}
    \label{eq:15}
    \left| \log\frac{x}{y} \right| &= \log\left( 1 + \frac{c}{a+b} \right) \\
                                   &\le \log\left( 1 + \frac{c}{a} \right)
                                   \le \log\left( 1 + \frac{\delta}{a} \right) .
  \end{align}
    Furthermore, 
  \begin{align}
    \log\left( \frac{a}{a-\delta} \right) - \log\left( 1 + \frac{\delta}{a} \right)
    & = \log \frac{1}{(a+\delta)(a-\delta)} \\
    & = \log \frac{1}{a^2 - \delta^2} \\
    & \ge \log \frac{1}{a^2} = - 2 \log a > 0.
  \end{align}
  
  \textbf{Case $x < y$.} Here, we can write $y = a + b$ and $x = y - c = a + b - c$ for $b \ge 0$ and $0 \le c \le \delta < a$.
  \begin{align}
    \left| \log\frac{x}{y} \right| &= \log\frac{y}{x} \\
                                   &= \log\left( \frac{a + b}{a+b-c} \right) \\
                                   &\le \log\left( \frac{a}{a-c} \right) \\
                                   &\le \log\left( \frac{a}{a-\delta} \right) = - \log\left( 1 - \frac{\delta}{a} \right) .
  \end{align}
\end{proof}

\begin{proof}[Proof of \cref{thm:confidence_bound}]
  We apply \cref{lem:p_phat} $N$ times and use the union bound to show that with probability greater than $1-\frac{\delta}{3}$ we have for every $n \in [N]$
  \begin{align}
    \label{eq:4}
    |p(X_n) - \htp{X_n}| \le \frac{K_{\mathrm{max}}}{w^d} \sqrt{\frac{\log\frac{6N}{\delta}}{2M}}
     + C_2 w .
  \end{align}

  Similarly, by \cref{lem:prob_p}, we have with probability greater than $1-\frac{\delta}{3}$ that
  \begin{align}
    \label{eq:9}
    p(X_n) \ge \frac{\delta}{3N}
  \end{align}
  for all $n \in [N]$.

  Again by the union bound, we have that with probability greater than $1- \frac{2\delta}{3}$ both \cref{eq:4} and \cref{eq:9} hold for all $n \in [N]$, and thus, by \cref{lem:log_ineq}, we obtain
    \begin{align}
    \label{eq:5}
    & \left| \widehat \Ent (\Dn; w) + \frac{1}{N} \sum_{n=1}^N \log p(X_n) \right| \\
    &\qquad= \left| \frac{1}{N} \sum_{n=1}^N \log \frac{p(X_n)}{\htp{X_n}} \right| \\
    &\qquad\le -\log \left( 1 - \frac{\frac{K_{\mathrm{max}}}{w^d} \sqrt{\frac{\log\frac{6N}{\delta}}{2M}} + C_2 w}{\frac{\delta}{3N}} \right) \\
    &\qquad= -\log \left( 1 - \frac{3 N K_{\mathrm{max}}}{w^d \delta} \sqrt{\frac{\log\frac{6N}{\delta}}{2M}} - \frac{3 N C_2 w}{\delta} \right) , \label{eq:empirical_sum}
  \end{align}
  provided the argument in the logarithm is positive.
  Finally, we have the upper bound on the variance
  \begin{align}
    \label{eq:2}
    \mathbb E &\left[\left(\Ent(X)  + \frac{1}{N} \sum_{n=1}^N \log p(X_n)\right)^2\right] \\ & = \frac{1}{N^2} \sum_{n=1}^N \Exp{(\Ent(X) + \log p(X))^2} \\
                                                                                      &= \frac{1}{N} (\Exp{\log^2 p(X)} - \Ent(X)^2) \\
                                                                                      &\le \frac{1}{N} C_1
  \end{align}
  and apply Chebychev's inequality, showing that with probability greater than $1 - \frac{\delta}{3}$,
  \begin{align}
        \left| \Ent(X) + \frac{1}{N} \sum_{n=1}^N \log p(X_n) \right| \le \sqrt{\frac{3 C_1}{N \delta}}. \label{eq:cheby}
  \end{align}
  The union bound and the triangle inequality applied to \cref{eq:empirical_sum,eq:cheby} yields the desired result.
\end{proof}

\section{Libraries Used}
For our experiments, we built upon code from the following sources.
\begin{itemize}[leftmargin=5pt]     \item VIBERT~\citep{mahabadi2021variational} at \texttt{\href{https://github.com/rabeehk/vibert}{github.com/rabeehk/vibert}}.
    \item TRANSFORMERS~\citep{wolf2019huggingface} at \texttt{\href{https://github.com/huggingface/transformers}{github.com/huggingface/transformers}}.
    \item {\doe}~\citep{pmlr-v108-mcallester20a} at \texttt{\href{https://github.com/karlstratos/doe}{github.com/karlstratos/doe}}.
    \item SMILE~\citep{song2019understanding} at \texttt{\href{https://github.com/ermongroup/smile-mi-estimator}{github.com/ermongroup/smile-mi-estimator}}.
    \item InfoNCE, MINE, NWJ, CLUB~\citep{Cheng2020CLUB} at \texttt{\href{https://github.com/Linear95/CLUB}{github.com/Linear95/CLUB}}.
\end{itemize}

\end{document}